\documentclass{article}

\PassOptionsToPackage{numbers, compress}{natbib}
 \usepackage[main, final]{neurips_2026}
\usepackage[utf8]{inputenc} % allow utf-8 input
\usepackage[T1]{fontenc}    % use 8-bit T1 fonts
\usepackage{hyperref}       % hyperlinks
\usepackage{url}            % simple URL typesetting
\usepackage{booktabs}       % professional-quality tables
\usepackage{amsfonts}       % blackboard math symbols
\usepackage{nicefrac}       % compact symbols for 1/2, etc.
\usepackage{microtype}      % microtypography
\usepackage{xcolor}         % colors

\usepackage{makecell}
\usepackage{bm}
\usepackage{amsmath}
\usepackage{graphicx}
\usepackage{wrapfig}
\usepackage{multirow}
\usepackage{subcaption}
\usepackage{enumitem}
\usepackage{multicol}
\usepackage{colortbl}
\definecolor{dark-gray}{gray}{0.30}

\usepackage{graphicx}
\usepackage{subcaption}
\usepackage{array}
\usepackage{amssymb}
\usepackage{algorithm}
\usepackage{algpseudocode}
\usepackage{listings}
\usepackage{xcolor}

\title{Learning Agent-Centric Vision-Language-Action World Models for Autonomous Driving}

\author{%
  Guoqing Wang \quad
  Pin Tang \quad
  Xiangxuan Ren \quad
  Chao Ma\thanks{Corresponding author.} \\
  MoE Key Lab of Artificial Intelligence, Institute of AI \\
  Shanghai Jiao Tong University, Shanghai, China \\
  \texttt{
  \{guoqing.wang,pin.tang,bunny\_renxiangxuan,chaoma\}@sjtu.edu.cn} \\[0.5em]
  \small Project page: \url{https://AgentWorld-AD.github.io}
}

\begin{document}

\maketitle

\begin{abstract}

Recent progress in integrating vision-language-action (VLA) models with world modeling has significantly advanced end-to-end autonomous driving by allowing systems to reason about imagined futures instead of merely reacting to current observations.
However, most world VLA models generate future frames at the pixel level with plenty of dense visual tokens.
Although such generated futures can be visually plausible, they preserve many driving-irrelevant appearance details, %while encoding planning-relevant factors, such as agent geometry, motion dynamics, and interactions, only implicitly, yet bring only limited performance gains. 
i.e., only a small portion of the tokens directly encode critical planning-relevant factors, such as agent geometry, motion dynamics, and interactions. This introduces a significant performance bottleneck. 
% As a result, existing world VLA models often require substantially more tokens than VLA base models, yet bring only limited performance gains.
An important question thus arises: \emph{what world state should a world VLA model reason over to support planning beyond pixel-level prediction?}
% However, future imagination relying solely on visual plausibility fails to provide the comprehensive information necessary for driving scene understanding tasks, such as planning.
% The main performance limitation arises from the interplay of ego-status extrapolation, language priors, dataset biases, and the behaviors of surrounding participants, all of which shape the envisioned future.
% This raises an important question: \emph{what is the most effective latent world state, beyond visual images, for reasoning?}
We argue that planning-oriented world modeling should focus on a compact and structured agent-centric latent state that captures the interactions between objects in the driving scene over time. 
As such, we introduce \textbf{AgentWorld}, an agent-centric VLA world model that uses discrete agent tokens jointly for reasoning and trajectory planning.
Each token is dynamically grounded to a surrounding agent associated with its object's state, such as semantic identity, 3D geometry, motion, and interaction context.
AgentWorld learns these tokens through three key processes: explicit geometric grounding, reasoning-oriented supervised fine-tuning, and agent-oriented reinforcement learning.
Extensive open-loop and closed-loop experiments under diverse driving scenarios show that AgentWorld enhances planning capabilities, %reduces reliance on shortcuts, 
fully exploits visual evidence and agent dynamics, and yields interpretable agent-centric world states. 

\end{abstract}

\section{Introduction} \label{sec:introduction}
The advancements in Vision-Language-Action (VLA)~\cite{zhou2025autovla, zhou2025opendrivevla, jiang2025alphadrive, ma2024dolphins, zhang2024chatscene, li2024llada, han2025dme, chi2025impromptu, arai2025covla, tang2026pixelpilot} and world models~\cite{wang2024driving, wang2023drivedreamer, wang2024worlddreamer, lu2025wovogen, jia2023adriver, min2024driveworld, hu2023gaia} have significantly propelled end-to-end autonomous driving~\cite{hu2023planning, jiang2023vad, chen2024vadv2, li2025semi, li2024ego}. 
VLA models inherit the instruction-following and reasoning capabilities of Multimodal Large Language Models (MLLMs)~\cite{achiam2023gpt4, bai2023qwenvl, bai2025qwen25vl, alayrac2022flamingo, liu2024llava, liu2024improved, ren2026grounding}, effectively connecting perception, reasoning, and action generation within a unified framework. 
World models complement this ability by predicting how the driving scene will evolve before an action is taken.
% allowing the policy to reason about possible futures instead of merely reacting to current observations. 
% Recent efforts aim to integrate these two approaches into world VLA models~\cite{wang2026learning, lu2025uniugp, cen2025worldvla, liu2026driveworld, zeng2025futuresightdrive}, enabling VLA systems to consider imagined futures to produce more reasonable actions. 
Recent efforts aim to integrate these two approaches into VLA world models~\cite{wang2026learning, lu2025uniugp, cen2025worldvla, liu2026driveworld, zeng2025futuresightdrive}, enabling VLA systems to consider imagined futures to produce more reasonable actions. 

\begin{figure}
    \centering
    \includegraphics[width=1\linewidth]{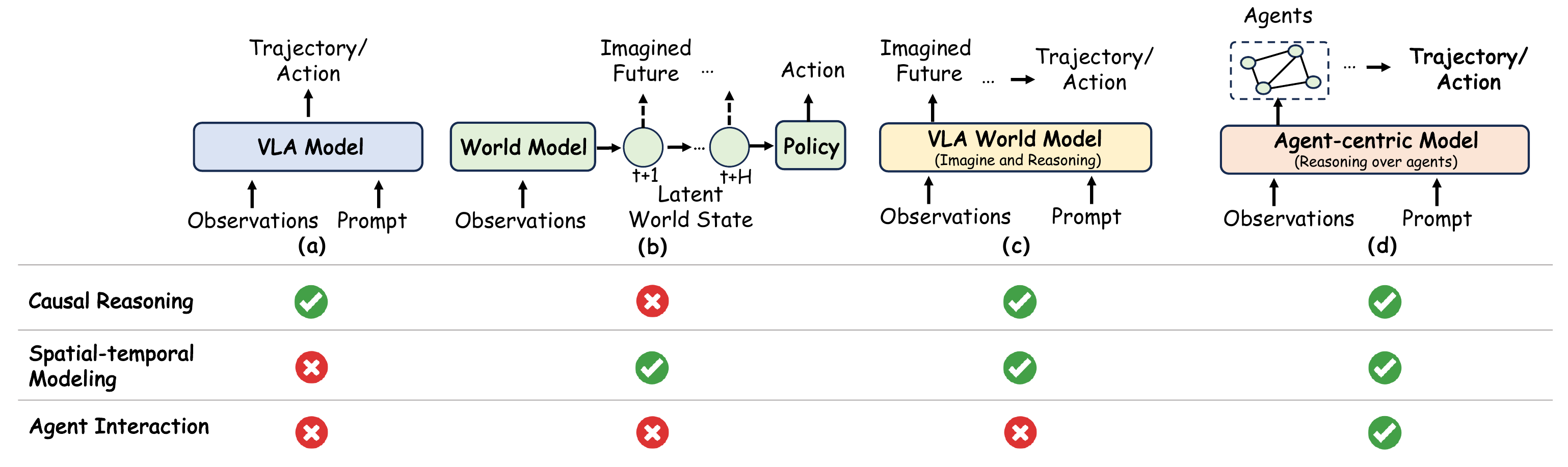}
    % \vspace{-15pt}
    \caption{(a) VLA models output actions from observations and prompts, lacking explicit spatiotemporal modeling and interaction awareness. (b) World models learn predictive latent futures, yet their imagined states fail to align with causal reasoning. (c) VLA World models integrate imagination and language reasoning, but reason over generated frames rather than interactive agents for driving decisions. (d) By contrast, AgentWorld models futures via discrete latent agent states, enabling VLA reasoning over physically grounded 3D geometry, motion dynamics and agent interactions. }
    \label{fig:paradigm}
    % \vspace{-18pt}
\end{figure}

% Despite this promising paradigm, most prior world VLA models realize future imagination through pixel-level observation generation with autoregressive or multi-step diffusion, leading to costly pixel-based world modeling and delayed policy decision-making. Moreover, future frames retain fine-grained visual details, including texture, lighting, background, and visual continuity, while agent physical states and interactions are encoded only implicitly.
% As a result, most visual tokens are occupied by appearance details that are irrelevant to planning, which not only increases reasoning overhead but also introduces distracting noise into the downstream policy. 
Despite this promising paradigm shown in Fig.~\ref{fig:paradigm}, most prior VLA world models imagine the future frames through pixel-level generation with autoregressive or multi-step diffusion. Although such generated future frames can be visually plausible, they retain numerous task-irrelevant visual details, i.e., texture, lighting, background, and visual continuity. 
Only a small portion of tokens explicitly encode
\begin{wrapfigure}{r}{0.46\textwidth}
% \vspace{-10pt}
\centering
    \includegraphics[width=1.0\linewidth]{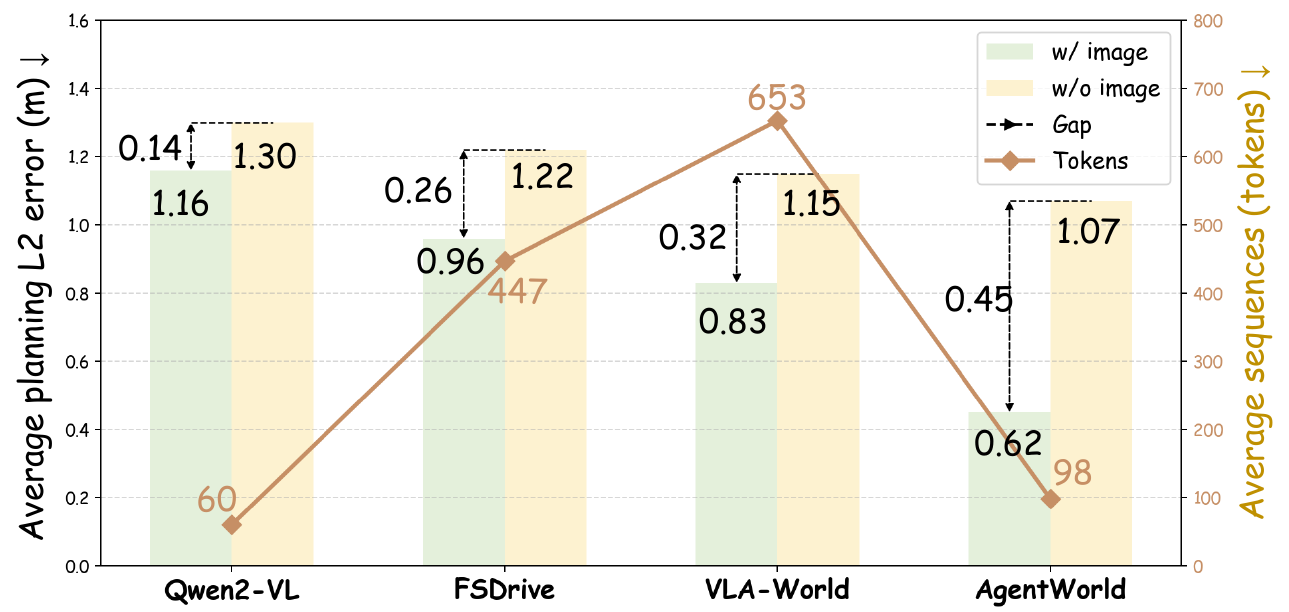}
    % \vspace{-10pt}
    \caption{Visual grounding results (UniAD metric) and token efficiency of VLA models.}
    % \vspace{-10pt}
    \label{fig:token}
\end{wrapfigure} 
planning-relevant cues, 
including agent geometry, motion dynamics, and interactive behaviors. This introduces a significant bottleneck in VLA world models. As illustrated in Fig.~\ref{fig:token}, existing VLA 
world models demand far more tokens than the fine-tuned base model, 
i.e., 
Qwen2-VL~\cite{wang2024qwen2vl}, yet bring only slight planning improvement. 
Moreover, the visual grounding ablation reveals that removing image cues only slightly degrades existing world VLA models, even with their heavy use of dense visual tokens.
This suggests that the visual inputs in these models are not yet fully transformed into planning-relevant reasoning, with many of the visual tokens instead being used for image generation.
% This demonstrates that these models fail to fully leverage visual inputs for planning. Instead, most visual computation focuses on appearance reconstruction, not task-driven reasoning.
As such, the core challenge is not simply how to generate visually plausible future frames, but how to integrate world modeling with VLA through a compact, structured, and task-aligned state. 
An important question thus arises: \textit{what world state should a VLA world model reason over to support planning beyond pixel-level prediction?}

Motivated by JEPA-style paradigms~\cite{lecun2022path, assran2023self, maes2026leworldmodel}, which prioritize predictive abstraction over pixel reconstruction, we argue that planning-oriented world modeling should emphasize a latent world state that captures the temporal interactions among objects in driving scenes. Rather than focusing on reconstructing dense visual futures, this state abstracts the scene into decision-relevant entities and their evolving relationships.
We implement this idea by breaking down the scene into temporally evolving agents, each defined by semantic identity, 3D geometry, motion state, and interaction context. This structured abstraction enables the policy to recognize relevant agents, predict their motion evolution, and reason about how their future behaviors will influence the ego’s trajectory. In this way, we align the latent world state with the causal structure of driving, shifting world modeling from future frame generation to structured reasoning about interacting agents.

In this paper, we introduce AgentWorld, an agent-based world VLA model designed for fully autonomous driving, as illustrated in Fig.~\ref{fig:paradigm}. AgentWorld integrates causal reasoning, predictive world modeling, and agent interactions into a cohesive framework. 
In AgentWorld, the driving environment is represented by a set of discrete agent tokens, labeled as \texttt{<agent\_i>}. These tokens are utilized for both reasoning and trajectory planning. 
% Each token corresponds to an agent slot that can either be dynamically linked to an active participant in the environment or designated as a "no-object" state. 
This structure creates a well-defined interface among perception, reasoning, and planning processes.
To make these latent agents physically meaningful, AgentWorld anchors them to 4D geometric features and 2D visual cues, and updates their hidden states through the MLLM into contextual representations of semantic identity, 3D geometry, motion, and interaction.
% These tokens are learned via explicit geometric grounding, reasoning-oriented supervised fine-tuning, and agent-oriented reinforcement learning with grouped relative policy optimization~\cite{guo2025deepseek}.
% As a result, AgentWorld provides a compact, physically grounded, and interaction-aware world state for interpretable planning over surrounding agents, while avoiding the excessive token cost and visual noise of pixel-level future imagination.
% Through the multimodal routed anchoring module, these tokens query offline 4D geometric features and 2D visual appearance cues. This allows each slot to capture relevant physical and semantic characteristics from the scene. The Multimodal Language and Learning Model (MLLM) further updates the hidden states at the \texttt{<agent\_i>} positions, transforming them into contextual agent representations that encapsulate semantic identity, 3D geometry, motion state, and interaction context.
%
AgentWorld learns these tokens via explicit geometric grounding, reasoning-oriented supervised fine-tuning (SFT), and agent-oriented reinforcement learning (RL) with group relative policy optimization~\cite{guo2025deepseek}. The resulting agent tokens deliver a compact, physically grounded, and interaction-aware representation of the world state, facilitating interpretable reasoning and planning regarding surrounding agents in driving scenarios.

Extensive open-loop and closed-loop experiments demonstrate the effectiveness of AgentWorld across diverse driving scenarios. 
By reasoning over interpretable agent-centric latent states, AgentWorld improves planning performance and better leverages visual evidence. 
These results validate agent-centric world states as an effective intermediate representation for VLA-based autonomous driving. 
Our contributions are summarized as follows:
\begin{itemize}[itemsep=1pt,topsep=0pt]
    \item We reveal that the bottleneck of existing world VLA models stems from the fact that %most visual tokens are used for image generation rather than planning-relevant factors, 
    a considerably large number of visual tokens for future generation are irrelevant to planning, 
    which motivates our design of compact, structured, planning-oriented latent world modeling.
    % We reveal that the bottleneck of existing world VLA models lies not merely in predicting plausible future frames, but in how world models are integrated with VLA policies.
    % We argue that VLA world models should reason beyond imagined visual futures, as planning requires structured causal states not fully specified by future generation alone, motivating a shift toward physically grounded agent abstractions.

    \item We introduce an \emph{agent-centric latent state} for planning-oriented world modeling, which represents the driving scene as temporally evolving agents grounded in 3D geometry, motion, semantics, and interaction context.

    \item We propose \textbf{AgentWorld}, an agent-centric world VLA model that instantiates this state with discrete agent tokens. Experiments on open-loop and closed-loop benchmarks show consistently improved planning performance and more interpretable world states. %reduced shortcut reliance, stronger use of surrounding-agent cues, and more interpretable world states.
\end{itemize}

\section{Related Work} \label{sec:related}
\subsection{Vision–Language–Action World Models}
VLA models~\cite{zhou2025autovla, zhou2025opendrivevla, tian2024drivevlm, zheng2025driveagent, mao2023language, xu2024drivegpt4, cen2025worldvla, qian2025agentthink, nie2024reason2drive, brohan2022rt, zitkovich2023rt, belkhale2024rt, black2024pi_0, driess2023palm, wang2026vggdrive,xie2026latentvla} and world models~\cite{wang2023drivedreamer, wang2024driving, brooks2024video, zheng2024genad, zheng2024doe, zheng2024occworld, zheng2025world4drive, yang2024generalized, wu2023mars, xu2025occ, yang2025resim, yang2025driving, li2024uniscene, li2025omninwm, gao2024vista, zhou2025hermes, bar2025navigation, assran2025v} have become two important directions for embodied intelligence and autonomous driving. Early driving VLMs~\cite{shao2024lmdrive, ma2024dolphins, sima2024drivelm, tian2024drivevlm, wang2025omnidrive, wang2024drivecot} demonstrate that language improves instruction following, scene understanding, and explicit reasoning in complex scenarios. However, language is often used as an auxiliary interface for explanation or high-level guidance, rather than being directly coupled with action generation.
Recent VLA methods~\cite{zhou2025opendrivevla, zhou2025autovla, jiang2025alphadrive} address this limitation by unifying multimodal perception, language reasoning, and trajectory prediction within an end-to-end policy.
In parallel, driving world models aim to improve planning through predictive scene evolution. Early works~\cite{hu2023gaia, wang2023drivedreamer, jia2023adriver, wang2024driving} generate future driving scenes conditioned on videos, actions, or text prompts, while subsequent studies~\cite{zhao2025drivedreamer, hu2024drivingworld, zheng2024occworld, yang2025driving,wang2026latent,liu2026driveva} extend this paradigm with multi-view generation, autoregressive spatiotemporal modeling, and 4D occupancy prediction. These methods suggest that autonomous driving can benefit from reasoning over possible future states instead of reacting only to current observations.
Recently, VLA world models have emerged by integrating VLA reasoning with world-model imagination. Representative approaches~\cite{wang2026learning, lu2025uniugp, liu2026driveworld, shang2026dynvla, luo2026unleashing,li2025drivevla,liu2026uni,tan2025latent, luo2026last} introduce future frames, latent world states, dynamic tokens, or spatiotemporal reasoning modules into action inference. Despite their progress, most existing methods still leave a fundamental question underexplored: \emph{what world state should a VLA world reason over to support planning beyond pixel-level prediction?} We argue that this question is central to driving, where decisions are governed by physically grounded multi-agent interactions.

\subsection{World State Representations for Autonomous Driving}
The above question is closely related to how world states are represented for planning.
Dense representations such as BEV~\cite{wang2025omnidrive, li2025end} and occupancy~\cite{liu2025occvla, dang2026sparseoccvla, liu2025drivepi} preserve 3D spatial structure and geometric details, but they remain low-level and often entangle traffic participants with background elements, making them inefficient interfaces for high-level policy reasoning.
Compact scene-level latents~\cite{zeng2025futuresightdrive, wang2026learning, shang2026dynvla,lin2025futurex} provide more efficient abstractions for future imagination, yet they typically compress multiple agents and interactions into global tokens, limiting structural interpretability and agent-level attribution.
Language-aligned world states~\cite{zhou2025autovla, nie2024reason2drive, mao2023language} naturally interface with MLLM reasoning, but are weakly grounded in physical quantities such as 3D geometry, motion, and inter-agent dynamics.
As a result, policies can still rely on ego-status extrapolation, language priors, or dataset-specific correlations instead of modeling the physical evolution of surrounding agents.
Consequently, existing representations either retain excessive low-level details, obscure agent-level causal factors, or lack explicit physical grounding, motivating a compact, interpretable, and agent-centric world state for driving-oriented reasoning.

\section{Method}
\begin{figure}
    \centering
    \includegraphics[width=1.0\linewidth]{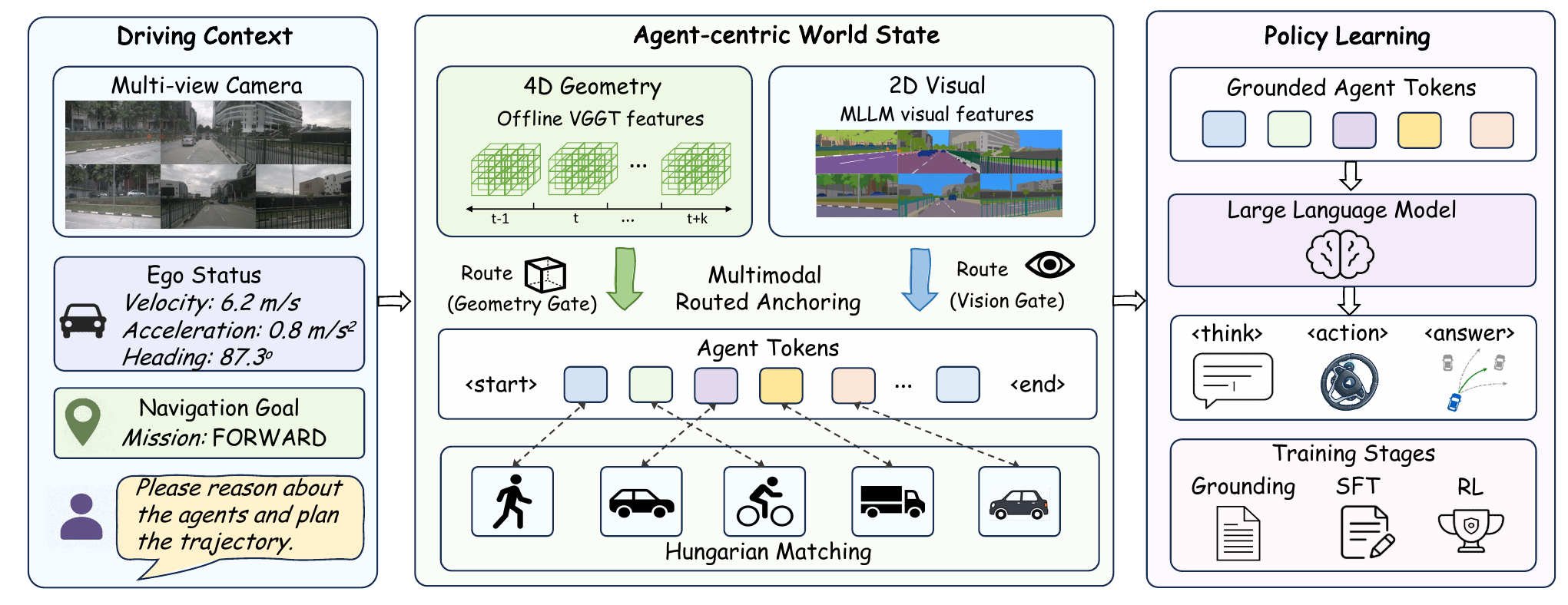}
    % \vspace{-10pt}
    \caption{Overview of AgentWorld. Given multi-view observations, ego-status, navigation goals, and language prompts, AgentWorld constructs an agent-centric world state by grounding discrete agent tokens with 4D geometry and 2D visual features through multimodal routed anchoring. The grounded agent tokens are then used by the policy model for reasoning, action, and planning through grounding, supervised fine-tuning, and reinforcement learning stages.}
    \label{fig:overview}
    % \vspace{-10pt}
\end{figure}
As illustrated in Fig.~\ref{fig:overview}, AgentWorld builds an agent-centric VLA world model for end-to-end autonomous driving. 
Instead of treating future images or global latent features as the reasoning state, AgentWorld represents the driving scene with discrete agent tokens that are grounded in 4D geometry, 2D visual features, and multimodal context. 
These grounded tokens act as structured world states, exposing surrounding participants and their interactions to the policy for reasoning and planning.
The method is organized into three components. 
Sec.~\ref{sec:rethinking_world_states} motivates agent-centric world states for VLA world models. 
Sec.~\ref{sec:grounding_agent_tokens} details how agent tokens are physically grounded through multimodal routed anchoring. 
Sec.~\ref{sec:policy_learning} describes the policy learning with grounded agent tokens.

% \begin{figure}
%     \centering
%     \includegraphics[width=1.0\linewidth]{figures/observations.pdf}
%     % \vspace{-15pt}
%     \caption{Empirical motivation for reasoning with agentic world state. (a) Removing ego-status causes smaller degradation in AgentWorld, suggesting less dependence on ego-motion shortcuts. (b) Adding visual inputs brings clearer gains to AgentWorld, indicating that visual evidence is more effectively grounded for planning. (c) The forward-dominated scene distribution in nuScenes highlights the risk of shortcut learning. Dashed lines denote absolute performance gaps (L2 metrics of UniAD~\cite{hu2023planning}).}
%     \label{fig:obs}
%     % \vspace{-20pt}
% \end{figure}
% \subsection{Problem Formulation: An Agentic World State}
\subsection{Rethinking World States for VLA World Models}
\label{sec:rethinking_world_states}
Let $\mathcal{H}_t=\{o_{1:t}, g\}$ denote the driving context at time $t$, where $o_{1:t}$ contains historical multi-view observations and ego-status, and $g$ is the navigation goal. 
The goal of end-to-end autonomous driving is to predict a future ego trajectory $\tau_{t:t+H}$. 
Following the VLA world model paradigm~\cite{wang2026learning, zeng2025futuresightdrive, shang2026dynvla}, planning can be viewed as decision-making augmented by predictive imagination:
\begin{equation}
    p(\tau_{t:t+H}, \hat{\mathcal{S}} \mid \mathcal{H}_t)
    =
    p_{w}(\hat{\mathcal{S}} \mid \mathcal{H}_t)
    \,
    p_{\pi}(\tau_{t:t+H} \mid \mathcal{H}_t, \hat{\mathcal{S}}).
\label{eq:vla_world_general}
\end{equation}
Here, $p_w$ denotes a world model that constructs an imagined world state $\hat{\mathcal{S}}$ from the driving context, while $p_\pi$ denotes a VLA that predicts the ego trajectory conditioned on both the current context and the imagined state. 
This abstraction reveals that the world state is not merely an auxiliary prediction target, but the information interface through which predictive imagination influences decision-making.

% As such, the effectiveness of a world VLA model depends critically on the form of state $\mathcal{S}$ on which the policy conditions its decisions.
% If this state mainly preserves visually plausible details, or compresses the scene into an entangled global latent, it cannot expose the physical variables that actually determine safe driving. 
% In such cases, the policy can bypass the imagined state and rely on easier shortcuts, such as ego-status extrapolation, language priors, or dataset-specific correlations. 
% Thus, the key question is not only how to imagine the future, but \emph{what latent world state should a world VLA model reason over beyond visual images?}
% As such, the effectiveness of a world VLA model depends critically on the form of state $\mathcal{S}$ used for policy reasoning.
% Pixel-level future observations require substantial computation to reconstruct dense appearance details, yet planning-relevant factors such as agent geometry, motion dynamics, and interactions remain implicit in pixels.
% Compact global latents improve efficiency, but often entangle multiple scene factors and obscure agent-level causal structure.

Therefore, the effectiveness of a VLA world model depends critically on the form of world state $\mathcal{S}$ over which the policy reasons. 
Pixel-level future observations provide an intuitive form of imagination, but require the policy to process many dense visual tokens, most of which are devoted to image generation rather than planning-relevant factors. 
As a result, agent geometry, motion dynamics, and interactions are only sparsely and implicitly encoded in pixels, leading to high computational costs and limited task alignment. 
Compact global latents~\cite{shang2026dynvla} improve efficiency, but often entangle heterogeneous scene factors and obscure the agent-level causal structure needed for planning.

% Consequently, the policy lacks a structured interface for planning-oriented reasoning and can fall back on shortcuts such as ego-status extrapolation, language priors, or dataset-specific correlations.
% This motivates our central design goal: to construct a latent world state beyond images that is compact, physically grounded, and directly aligned with driving decisions.

% We empirically observe that existing VLA and VLA-world models indeed exhibit shortcut-driven behavior. 
% As shown in Fig.~\ref{fig:obs}(a), removing ego-status causes a substantial performance drop for existing baselines, suggesting that their planning results rely heavily on historical ego-motion priors such as velocity and heading, rather than sufficiently modeling surrounding-agent dynamics. 
% Fig.~\ref{fig:obs}(b) provides a complementary view: adding visual observations yields only limited gains, indicating that raw visual information or image-level future predictions are not effectively transformed into decision-useful physical knowledge. 
% This issue is further amplified by dataset bias, as shown in Fig.~\ref{fig:obs}(c), where straight-driving scenes dominate and allow policies to achieve reasonable average performance through ego-status extrapolation and common driving priors. 
% In contrast, AgentWorld exhibits a smaller ego-status gap and benefits more clearly from visual inputs, suggesting that grounding visual and geometric evidence into agent-level tokens provides a more effective reasoning interface.

This motivates an agent-centric world state. 
For driving, the planning-relevant future is largely shaped by the evolution of surrounding participants: their 3D geometry, motion states, and interactions with the ego vehicle. 
Rather than representing the world state as dense pixels, language-only descriptions, or a compact scene latent, AgentWorld organizes the scene around potential agents. 
Specifically, we introduce a fixed set of special agent tokens into the MLLM sequence,
\begin{equation}
    \mathcal{T}_{\mathrm{agent}}
    =
    \{
    \langle \mathrm{agent}\_1 \rangle,
    \langle \mathrm{agent}\_2 \rangle,
    \cdots,
    \langle \mathrm{agent}\_N \rangle
    \},
\end{equation}
where $N$ is the maximum number of agent tokens. After being grounded with multimodal evidence and updated by the MLLM, the hidden states at these token positions define the agent-centric latent state:
\begin{equation}
    \mathcal{S}_t^{\mathrm{agent}}
    =
    \{
    \mathbf{h}_{t}^{1},
    \mathbf{h}_{t}^{2},
    \cdots,
    \mathbf{h}_{t}^{N}
    \},
    \quad
    \mathbf{h}_{t}^{i}\in\mathbb{R}^{D_{\mathrm{LLM}}},
\label{eq:agent_world_state}
\end{equation}
where $\mathbf{h}_{t}^{i}$ is the MLLM hidden state at the position of $\langle \mathrm{agent}\_i \rangle$. 
Each token acts as an addressable agent slot that can be dynamically grounded to a surrounding participant or assigned to a "no-object" state. 
Through explicit physical grounding, active tokens are encouraged to encode semantic identity, 3D geometry, motion, and interaction context.

Under this formulation, AgentWorld performs planning by reasoning over grounded agent tokens:
\begin{equation}
    p(\tau_{t:t+H} \mid \mathcal{H}_t)
    =
    p_{\pi}
    \left(
    \tau_{t:t+H}
    \mid
    \mathcal{H}_t,
    \mathcal{S}_t^{\mathrm{agent}}
    \right).
\label{eq:agent_policy}
\end{equation}
The role of the world model is to construct $\mathcal{S}_t^{\mathrm{agent}}$ as a compact, physically grounded, and interaction-aware state, rather than reconstructing all future visual details. 
% Future information is injected into this interface through agent-level motion and trajectory grounding, so that the policy can reason over how surrounding participants may evolve and affect the ego plan. 
These tokens are trained with semantic identity, 3D geometry, motion, and interaction context, encouraging the policy to infer how surrounding participants are likely to evolve and how their dynamics may influence the ego vehicle.
This shifts VLA world models from future frame imagination toward agent-oriented reasoning.

\subsection{Grounding Agent Tokens as World States}
\label{sec:grounding_agent_tokens}
% To model agent-level world evolution, each token in $\mathcal{S}_t$ should be tied to a real traffic participant. 
% However, adding latent tokens alone does not ensure this correspondence. 
% Without explicit grounding, the tokens may degenerate into global scene embeddings or capture shortcut cues unrelated to verifiable agents. 
% We therefore construct agentic world states in three steps: agent token initialization, multimodal routed anchoring, and explicit agent-level supervision.
Given the agent-centric world state defined in Sec.~\ref{sec:rethinking_world_states}, we now describe how each agent token is grounded to a physically verifiable participant. The grounding process consists of three steps: agent token initialization, multimodal routed anchoring, and physical grounding objective.

\paragraph{Agent token initialization.} We initialize agent queries from the learnable embeddings of the special agent tokens, rather than introducing a separate object-query set. 
Let
\begin{equation}
    \mathbf{E}_{\mathrm{agent}}
    =
    \mathrm{Emb}(\mathcal{T}_{\mathrm{agent}})
    \in \mathbb{R}^{N\times D_{\mathrm{LLM}}},
\end{equation}
where each slot corresponds to one agent token. 
The initial agent queries are obtained by
\begin{equation}
    \mathbf{Q}_{t}^{0}
    =
    \mathbf{E}_{\mathrm{agent}}\mathbf{W}_{q}
    =
    [\mathbf{q}_{t}^{1,0};\cdots;\mathbf{q}_{t}^{N,0}]
    \in \mathbb{R}^{N\times d}.
\end{equation}
Thus, each special token is itself a learnable agent query. 
At initialization, these queries are ungrounded latent slots that define where agent-centric world states will be formed in the MLLM sequence, without assuming any fixed correspondence to physical agents.

\begin{wrapfigure}{r}{0.54\textwidth}
\centering
    \includegraphics[width=1.0\linewidth]{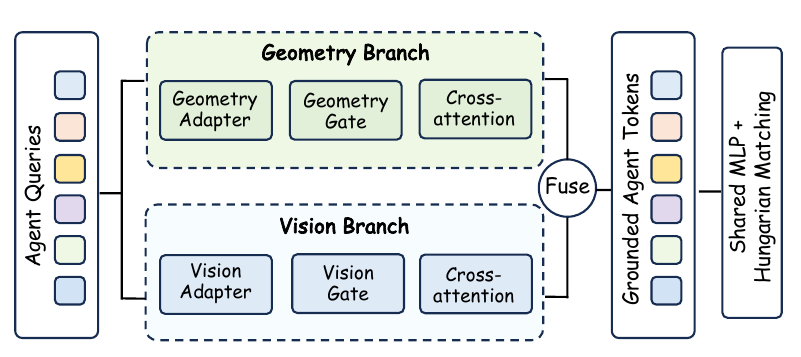}
    % \vspace{-10pt}
    \caption{The pipeline of the multimodal routed anchoring. Agent queries are routed to geometry and vision branches, and fused through cross-attention to produce grounded agent tokens.}
    \label{fig:route} 
    % \vspace{-5pt}
\end{wrapfigure}
\paragraph{Multimodal routed anchoring.}
The architecture of the multimodal routed anchoring module is shown in Fig.~\ref{fig:route}.
To ground the initialized agent queries, AgentWorld routes each slot to two complementary features: 4D geometry $\mathbf{G}^{0}_{t}$ from offline VGGT-style features following~\cite{wang2025vggt, chen2025think, zheng2025learning} and 2D visual $\mathbf{V}^{0}_{t}$ from the MLLM visual encoder. The geometry branch provides spatial layout, temporal consistency, and motion-related cues, while the vision branch provides appearance and semantic context. 
Both features are projected into a shared routing space:
\begin{equation}
    \mathbf{G}_{t}=\phi_g(\mathbf{G}^{0}_{t}),
    \quad
    \mathbf{V}_{t}=\phi_v(\mathbf{V}^{0}_{t}).
\end{equation}

For each routing layer, an agent query is processed by two parallel branches. 
Each branch first adapts the query to its modality, then uses a token-wise gate to modulate the query before cross-attention:
\begin{equation}
    \mathbf{R}_{m}^{\ell}
    =
    \mathrm{CrossAttn}_{m}
    \left(
    \boldsymbol{\gamma}_{m}^{\ell}\odot \psi_m(\mathbf{Q}_{t}^{\ell}),
    \phi_m(\mathbf{M}_{m}),
    \phi_m(\mathbf{M}_{m})
    \right),
    \quad
    \boldsymbol{\gamma}_{m}^{\ell}
    =
    \sigma(f_m(\mathbf{Q}_{t}^{\ell})),
\end{equation}
where $m\in\{g,v\}$, $\mathbf{M}_{g}=\mathbf{G}_{t}$, and $\mathbf{M}_{v}=\mathbf{V}_{t}$. 
The independent gates allow each token to decide how much geometry and vision to use, without forcing the two modalities to compete.

The geometry and vision responses are fused to update the agent queries:
\begin{equation}
    \mathbf{Q}_{t}^{\ell+1}
    =
    \mathrm{Fuse}
    \left(
    \mathbf{Q}_{t}^{\ell},
    \mathbf{R}_{g}^{\ell},
    \mathbf{R}_{v}^{\ell}
    \right).
\end{equation}
After $L$ routing layers, the resulting grounded queries $\tilde{\mathbf{Q}}_{t}$ are projected to the MLLM hidden space and inserted into the corresponding $\langle \mathrm{agent}\_i\rangle$ positions. 
The MLLM then contextualizes these tokens, and the hidden states at the corresponding positions serve as the grounded latent states $\{\mathbf{h}_{t}^{i}\}_{i=1}^{N}$. 
These states are subsequently supervised by the shared agent head and Hungarian matching.

\paragraph{Physical grounding objective.}
We ground each agent token by supervising its MLLM hidden state $\mathbf{h}_t^i$ with a lightweight training-time head:
\begin{equation}
    h_{\mathrm{agent}}(\mathbf{h}_t^i)
    =(\hat{\mathbf{p}}_t^i,\hat{\mathbf{b}}_t^i,\hat{\boldsymbol{\tau}}_t^i),
\end{equation}
which predicts category/existence, 3D bbox, and future trajectory. 
Because token order is unordered, predicted slots are matched to ground-truth agents with DETR-style bipartite matching~\cite{carion2020end}; matched slots receive classification, box, and trajectory losses, while unmatched slots are trained as no-object. 
This training-time head makes active tokens physically verifiable, leaving the grounded hidden states as agent-centric world states.

% Although modality routing injects scene evidence into agent tokens, the tokens remain unconstrained without direct physical supervision. 
% We therefore apply a shared lightweight MLP head to all routed tokens and predict an agent-level state for each slot:
% $\hat{\mathbf{y}}_{t}^{i}=h_{\mathrm{agent}}(\mathbf{q}_{t}^{i})=(\hat{\mathbf{c}}_{t}^{i},\hat{\mathbf{b}}_{t}^{i},\hat{\mathbf{v}}_{t}^{i})$, 
% $\hat{\mathbf{y}}_{t}^{i}=h_{\mathrm{agent}}(\mathbf{q}_{t}^{i})=(\hat{\mathbf{c}}_{t}^{i},\hat{\mathbf{b}}_{t}^{i},\hat{\mathbf{v}}_{t}^{i})$, including category or existence, 3D physical state, and motion state. 
% Since the token order is not predefined, we use DETR-style bipartite matching~\cite{carion2020end} to associate predicted tokens with ground-truth agents:
% \begin{equation}
%     \sigma^{\star}
%     =
%     \arg\min_{\sigma}
%     \sum_{j=1}^{M}
%     \mathcal{C}
%     \left(
%     h_{\mathrm{agent}}(\mathbf{q}_{t}^{\sigma(j)}),
%     \mathbf{y}_{t}^{j}
%     \right),
% \end{equation}
% where $\mathcal{C}$ combines classification, physical-state, and motion costs. 
% Matched tokens are supervised to regress the corresponding agent states, while unmatched tokens are treated as empty slots. 
% This grounding objective makes each active token verifiable through physical quantities and prevents the agent-centric world state from degenerating into free latent variables.

\subsection{Policy Learning with Agent Tokens}
\label{sec:policy_learning}
Grounded agent tokens serve as the world-state interface for policy learning. 
AgentWorld exposes the learned slots through the \texttt{<perception>} field,
\[
\texttt{<perception>}
\langle \mathrm{agent}\_1\rangle \cdots \langle \mathrm{agent}\_N\rangle
\texttt{</perception>},
\]
where the state of each participant is carried by the token hidden state rather than an explicit text description. 
The policy is trained to output a structured response $y=(\xi,a,\tau)$, consisting of a concise agent-grounded rationale \texttt{<think>}, a high-level driving action \texttt{<action>}, and future ego waypoints \texttt{<answer>}. 
This format forces planning to be expressed through addressable participants instead of dense perception text or image-level future prediction.

\paragraph{Reasoning-oriented SFT.}
After the physical grounding stage, we train the MLLM to use agent tokens as the interface for planning. 
Each expert response is organized as $y^\star=(\xi^\star,a^\star,\tau^\star)$, where $\xi^\star$ is a concise rationale over relevant agent tokens, $a^\star$ is a high-level driving action, and $\tau^\star$ is the expert future trajectory. 
We optimize the standard autoregressive objective:
\begin{equation}
    \mathcal{L}_{\mathrm{SFT}}
    =
    -
    \sum_{k}
    \log p_{\theta}
    \left(
    y_k^\star
    \mid
    y_{<k}^\star,
    \mathcal{H}_t,
    \mathcal{S}^{\mathrm{agent}}_t
    \right),
    \quad
    \mathcal{S}^{\mathrm{agent}}_t=\{\mathbf{h}_t^i\}_{i=1}^{N}.
\end{equation}
The rationale $\xi^\star$ is deliberately compact: it does not verbalize all perception results, but only identifies planning-relevant agent tokens, summarizes their interaction risk, and explains the resulting maneuver. 
For example, the model may state that \texttt{<agent\_3>} is crossing the ego path and therefore the ego vehicle should decelerate and keep lane. 
This supervision encourages the MLLM to bind its reasoning and action generation to the grounded latent agent-centric world states rather than relying solely on ego-status history or generic driving priors. 
To bridge training and inference, the final SFT phase adopts the same MLLM-only input-output format used in RL, where the auxiliary grounding heads and routed anchoring modules are no longer required in inference.

\paragraph{Agent-oriented RL.}
We further optimize AgentWorld with GRPO~\cite{guo2025deepseek}. 
Starting from the reasoning-oriented SFT checkpoint, the model samples multiple structured responses for each scene, each containing a concise rationale, a high-level action, and future ego waypoints. 
These candidates are evaluated by a composite reward that measures format, perception, action, trajectory, and interaction context. 
The trajectory reward encourages smoothness and imitation of expert behavior following~\cite{wang2026learning}, while the interaction reward explicitly evaluates the ego rollout against surrounding agents and penalizes collisions, insufficient clearance, and short time-to-collision. 
This reward design makes the RL stage agent-oriented: improving the final trajectory requires the policy to account for the dynamic agents represented by the learned \texttt{<agent\_i>} tokens. 

GRPO ranks multiple responses sampled from the same scene by their relative rewards, using the SFT model as a reference to stabilize updates. 
During RL, only the MLLM policy is updated. 
The auxiliary agent head is retained solely as a frozen reward generator for agent-token consistency and interaction evaluation. At inference time, the reward-side agent head is not required and can be discarded.
Thus, RL improves driving behavior while preserving the agent-token reasoning interface learned during SFT. 
Detailed rewards are provided in the appendix.

\section{Experiments}
\label{sec:exp}

\begin{table*}[t]
    \centering
    \setlength{\tabcolsep}{1.0mm}
    \caption{End-to-end trajectory planning results on nuScenes~\cite{caesar2020nuscenes}. We evaluate L2 error and collision rate following the respective protocols of ST-P3~\cite{hu2022stp3} and UniAD~\cite{hu2023planning}. The $*$ indicates the use of additional ego-state information.}
    % \vspace{-5pt}
    % \resizebox{\columnwidth}{!}{
    \resizebox{1.0\linewidth}{!}{
    \begin{tabular}{lcccccccc|cccccccc|c}
    \toprule
     \multirow{4}{*}{\textbf{Method}} & \multicolumn{8}{c}{\textbf{ST-P3 metrics}} & \multicolumn{8}{c}{\textbf{UniAD metrics}} & \multirow{4}{*}{{\textbf{VLM}}}  \\
    \cmidrule(lr){2-9} \cmidrule(lr){10-17}
    & \multicolumn{4}{c}{\textbf{L2 (m)} $\downarrow$} & \multicolumn{4}{c}{\textbf{Collision (\%)} $\downarrow$} & \multicolumn{4}{c}{\textbf{L2 (m)} $\downarrow$} & \multicolumn{4}{c}{\textbf{Collision (\%)} $\downarrow$} &   \\
    \cmidrule(lr){2-5} \cmidrule(lr){6-9} \cmidrule(lr){10-13} \cmidrule(lr){14-17}
    & 1s & 2s & 3s & \cellcolor{gray!20}Avg. & 1s & 2s & 3s & \cellcolor{gray!20}Avg. & 1s & 2s & 3s & \cellcolor{gray!20}Avg. & 1s & 2s & 3s & \cellcolor{gray!20}Avg. &  \\
    \midrule
    % \multicolumn{18}{c}{\textbf{Non-Autoregressive methods}} \\
    % \midrule
    % ST-P3*~\cite{hu2022stp3}& 1.33 & 2.11 & 2.90 &   \cellcolor{gray!30}2.11 & 0.23 & 0.62 & 1.27 &   \cellcolor{gray!30}0.71 & - & - & - &   - & - & - & - &   - & -  \\ 
    % % 628.3

    %  VAD~\cite{jiang2023vad} & 0.69 & 1.22 & 1.83 &   \cellcolor{gray!30}1.25 & 0.06 & 0.68 & 2.52 &   \cellcolor{gray!30}1.09 & - & - & - &   - & - & - & - &   - & -  \\
    % VAD*~\cite{jiang2023vad} & 0.17 & 0.34 & 0.60 &   \cellcolor{gray!30}0.37 & 0.04 & 0.27 & 0.67 &   \cellcolor{gray!30}0.33 & - & - & - &   - & - & - & - &   - & -  \\
    
    % UniAD~\cite{hu2023planning} & - & - & - & - & - & - & - & - & 0.59 & 1.01 & 1.48 &   \cellcolor{gray!30}1.03 & 0.16 & 0.51 & 1.64 &   \cellcolor{gray!30}0.77 & -  \\
    % UniAD*~\cite{hu2023planning} & - & - & - & - & - & - & - & - & 0.20 & 0.42 & \textbf{0.75} &   \cellcolor{gray!30}0.46 & 0.02 & 0.25 & 0.84 &   \cellcolor{gray!30}0.37 & -  \\

    % BEV-Planner~\cite{li2024ego} & 0.30 & 0.52 & 0.83 &   \cellcolor{gray!30}0.55 & 0.10 & 0.37 & 1.30 &   \cellcolor{gray!30}0.59 & - & - & - &   - & - & - & - &   - & -  \\
    % BEV-Planner*~\cite{li2024ego} & 0.16 & 0.32 & 0.57 &   \cellcolor{gray!30}0.35 & \textbf{0.00} & 0.29 & 0.73 &   \cellcolor{gray!30}0.34 & - & - & - &   - & - & - & - &   - & -  \\

    % PreWorld~\cite{li2025semi} & - & - & - &   - & - & - & - & - & 0.49  & 1.22 & 2.32 & \cellcolor{gray!30}1.34 & 0.19 & 0.57 & 2.65 & \cellcolor{gray!30}1.14&- \\
    
    % \midrule
    % \multicolumn{18}{c}{\textbf{Autoregressive methods}} \\
    % \midrule

     ELM~\cite{zhou2024embodied}& - & - & - &   - & - & - & - & -&  0.34  & 1.23 & 2.57 & \cellcolor{gray!30}1.38 & 0.12 & 0.50 & 2.36 & \cellcolor{gray!30}0.99 & BLIP2-2.7B\\
    FeD*~\cite{zhang2023coaching} & - & - & - &   - & - & - & - & -& 0.27  & 0.53 & 0.94 & \cellcolor{gray!30}0.58 & \textbf{0.00} & \textbf{0.04} & 0.52 & \cellcolor{gray!30}0.19 & LLaVA-7B\\

    OccWorld~\cite{zheng2024occworld} & 0.39  & 0.73 & 1.18 & \cellcolor{gray!30}0.77 & 0.11 & 0.19 & 0.67 & \cellcolor{gray!30}0.32 & 0.52  & 1.27 & 2.41 & \cellcolor{gray!30}1.40 & 0.12 & 0.40 & 2.08 & \cellcolor{gray!30}0.87 &GPT3-like\\
    
    Doe-1~\cite{zheng2024doe}  & 0.37  & 0.67 & 1.07 & \cellcolor{gray!30}0.70 & 0.02 & 0.14 & 0.47 & \cellcolor{gray!30}0.21& 0.50  & 1.18 & 2.11 & \cellcolor{gray!30}1.26 & 0.04 & 0.37 & 1.19 & \cellcolor{gray!30}0.53 &Lumina-mGPT-7B\\
    
    RDA-Driver*~\cite{RDA-Driver}& 0.17 & 0.37 & 0.69 &   \cellcolor{gray!30}0.40 & 0.01 & 0.05 & 0.26 &   \cellcolor{gray!30}0.10 & 0.23 & 0.73 & 1.54 &   \cellcolor{gray!30}0.83 & \textbf{0.00} & 0.13 & 0.83 &   \cellcolor{gray!30}0.32 & LLaVA-7B  \\
    EMMA*~\cite{hwang2024emma} & 0.14 & 0.29 & 0.54 & \cellcolor{gray!30}0.32 & - & - & - &   - & - & - & - & - & - & - & - & - & Gemini 1.0 Nano-1 \\
    % OmniDrive~\cite{wang2025omnidrive} & 0.40  & 0.80 & 1.32 & \cellcolor{gray!30}0.84 & 0.04 & 0.46 & 2.32 & \cellcolor{gray!30}0.94 & - & - &  - & - & - & - & - & - & LLaVA-7B \\
    OmniDrive*~\cite{wang2025omnidrive} & 0.14  & 0.29 & 0.55 & \cellcolor{gray!30}0.33 & \textbf{0.00} & 0.13 & 0.78 & \cellcolor{gray!30}0.30  & - & - & - & - & - & - & - & - & LLaVA-7B \\
    
    % FSDrive~\cite{zeng2025futuresightdrive} & 0.28  & 0.52 & 0.80 & \cellcolor{gray!30}0.53 & 0.06 & 0.13 & 0.32 & \cellcolor{gray!30}0.17 & 0.40  & 0.89 & 1.60 & \cellcolor{gray!30}0.96 & 0.07 & 0.12 & 1.02 & \cellcolor{gray!30}0.40& Qwen2-VL-2B  \\
    FSDrive*~\cite{zeng2025futuresightdrive} & 0.14  & 0.25 & 0.46 & \cellcolor{gray!30}0.28 & 0.03 & 0.06 & 0.21 & \cellcolor{gray!30}0.10 & 0.18  & 0.39 & 0.77 & \cellcolor{gray!30}0.45 & \textbf{0.00} & 0.06 & 0.42 & \cellcolor{gray!30}0.16 & Qwen2-VL-2B  \\
    % \midrule
    % VLA-World~\cite{wang2026learning} & 0.11 & 0.27 & 0.52 & \cellcolor{gray!30}0.30 & \textbf{0.00} & \textbf{0.03} & 0.26 & \cellcolor{gray!30}0.10  &  0.38 & 0.74 & 1.38 & \cellcolor{gray!30}0.83 & 0.02 & 0.08 & 0.36 & \cellcolor{gray!30}0.16& Qwen2-VL-2B  \\
    VLA-World*~\cite{wang2026learning} & 0.10 & 0.24 & 0.45 & \cellcolor{gray!30}0.26 & 0.02 & 0.05 & 0.18 & \cellcolor{gray!30}0.08 & 0.10 & 0.35 & 0.80 & \cellcolor{gray!30}0.42 & 0.01 & 0.05 & 0.30 & \cellcolor{gray!30}0.12& Qwen2-VL-2B  \\
    \midrule
    \rowcolor{gray!15}\textbf{AgentWorld} & 0.10 & 0.25 & 0.47 & \cellcolor{gray!30}0.27 & 0.01 & 0.04 & 0.18 & \cellcolor{gray!30}0.08 & 0.13 & 0.38 & 0.85 & \cellcolor{gray!30}0.45 & 0.01 & 0.05 & 0.31 & \cellcolor{gray!30}0.12 & Qwen2-VL-2B \\
    \rowcolor{gray!15}\textbf{AgentWorld*} & \textbf{0.09} & \textbf{0.23} & \textbf{0.44} & \cellcolor{gray!30}\textbf{0.25} & 0.01 & \textbf{0.03} & \textbf{0.17} & \cellcolor{gray!30}\textbf{0.07} & \textbf{0.09} & \textbf{0.34} & \textbf{0.77} & \cellcolor{gray!30}\textbf{0.40} & 0.01 & \textbf{0.04} & \textbf{0.28} & \cellcolor{gray!30}\textbf{0.11} & Qwen2-VL-2B  \\
    \bottomrule
    \end{tabular}
    }
    % \vspace{-5pt}
    % }
    \label{tab:plan}
\end{table*}

\subsection{Experimental Setups}
\paragraph{Datasets and metrics.}
We evaluate AgentWorld on nuScenes~\cite{caesar2020nuscenes} and Bench2Drive~\cite{jia2024bench2drive}, covering both open-loop planning and closed-loop driving. 
For nuScenes, we report standard trajectory L2 error and collision rate. 
For Bench2Drive, we follow the official protocol and report Driving Score, Success Rate, Efficiency, Comfortness when available. 
More details of metrics can be found in the appendix.
% We evaluate AgentWorld on both open-loop and closed-loop autonomous driving benchmarks. 
% For open-loop planning, we use nuScenes~\cite{caesar2020nuscenes}, where the model takes historical multi-view observations, ego-status, and navigation commands as input and predicts future ego waypoints. 
% We report standard planning metrics, including trajectory L2 error at different horizons and collision rate, and further conduct diagnostic evaluations under ego-status removal and visual ablation to measure shortcut reliance and visual grounding. 
% For closed-loop evaluation, we use Bench2Drive~\cite{jia2024bench2drive}, where the predicted actions are executed in simulation and evaluated under interactive traffic scenarios. 
% We report the official metrics, including Driving Score, Route Completion, and Infraction Score, together with fine-grained safety indicators such as collisions, red-light violations, lane invasions, and off-road behaviors when available. 
% These benchmarks jointly evaluate imitation accuracy, shortcut robustness, and closed-loop interaction safety. 
% More details about dataset splits, evaluation protocols, and metric definitions are provided in the supplementary material.

\paragraph{Implementation details.} 
AgentWorld is initialized from Qwen2-VL and Qwen2.5-VL series models~\cite{wang2024qwen2vl,bai2025qwen25vl}, with multi-view images as visual inputs and frozen VGGT-style features as offline 4D geometry features. We train the model in three stages: physical grounding SFT, reasoning-oriented SFT followed by agent-oriented RL. The SFT stages jointly supervise grounded agent tokens, high-level actions, reasoning traces, and future waypoints, while the RL stage optimizes sampled responses using format, perception, action, trajectory, and agent-interaction rewards. All experiments are implemented in PyTorch on $4\times80$GB NVIDIA A800 GPUs. Detailed hyperparameters, prompts, reward designs, and training schedules are provided in the appendix.
% We initialize AgentWorld from Qwen2-VL~\cite{wang2024qwen2vl} and Qwen2.5-VL~\cite{bai2025qwen25vl} series models following recent VLA driving models~\cite{wang2026learning, zeng2025futuresightdrive}. 
% Historical multi-view images are used as visual inputs, while a frozen VGGT-style encoder extracts offline 4D geometry features to construct the geometry memory bank. We train the model in two stages.
% In the first stage, we perform agent-aware supervised fine-tuning, where the model learns to ground agent tokens with category, 3D state, velocity, and short-horizon motion supervision, and meanwhile predicts the reasoning trace, high-level action, and future waypoints. 
% This stage is optimized with AdamW for 12 epochs using an initial learning rate of $1\times10^{-4}$. In the second stage, starting from the SFT checkpoint, we further optimize the policy with GRPO for one epoch. For each prompt, we sample $K=8$ candidate responses and compute rewards from output format, high-level action correctness, trajectory quality, and agent token reference. 
% The GRPO stage uses a learning rate of $1\times10^{-6}$, a global batch size of 8, and a KL regularization term to stabilize optimization with respect to the SFT reference policy. 
% All experiments are implemented in PyTorch and trained on $4\times80$GB Nvidia A800 GPUs. 
% More detailed hyperparameters, prompt templates, reward definitions, and training schedules are provided in the supplementary material.

\subsection{Main Results}
\begin{figure}[t]
    \centering
    \includegraphics[width=1.0\linewidth]{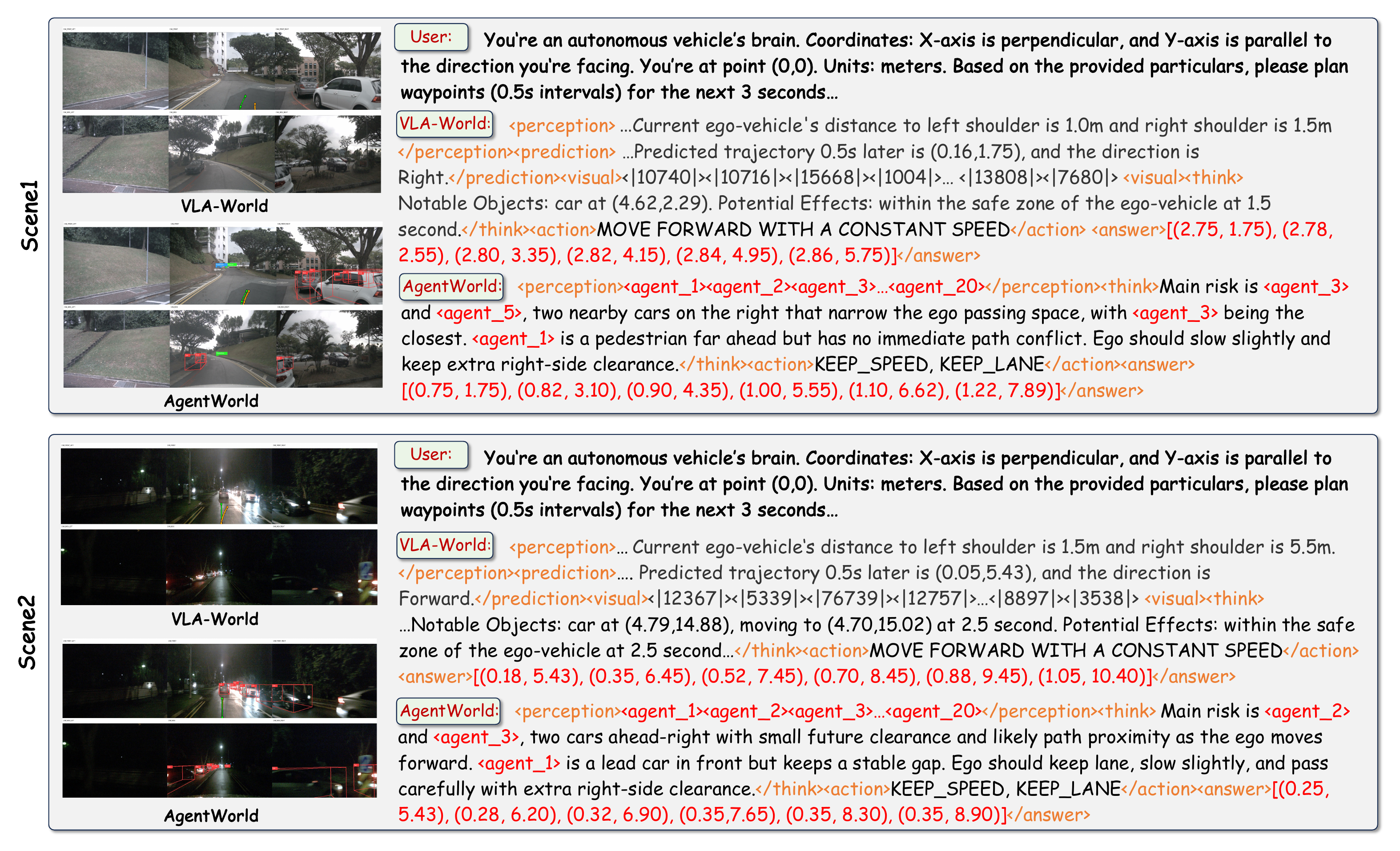}
    % \vspace{-10pt}
    \caption{Qualitative comparison between VLA-World~\cite{wang2026learning} and AgentWorld. For each scene, the left panels visualize the future ego trajectories of VLA-World (top) and AgentWorld (bottom), with AgentWorld additionally showing the traffic participants aligned with the special tokens $\langle \mathrm{agent}\_i\rangle$. The right panels present the corresponding perception, reasoning, and planning results.}
    \label{fig:vis_com}
    % \vspace{-10pt}
\end{figure}

\begin{table}[t]
    \scriptsize
    \centering
    \begin{minipage}[t]{0.44\linewidth}
        % \vspace{-5pt}
        \centering
        \makeatletter\def\@captype{table}
        \setlength{\tabcolsep}{3mm}
        \caption{Perception performance on nuScenes val set. We evaluate 3D object detection using both mAP and NDS metrics.}
        \label{tab:perception}
        % \vspace{3pt}
        \begin{tabular}{lcc}
        \toprule
        \textbf{Method} 
        & \textbf{mAP} $\uparrow$ 
        & \textbf{NDS} $\uparrow$ \\
        \midrule
        CenterNet~\cite{zhou2019objects} & 0.306 & 0.328 \\
        FCOS3D~\cite{wang2021fcos3d} & 0.299 & 0.373 \\
        DETR3D~\cite{wang2022detr3d} & 0.303 & 0.374 \\
        PGD~\cite{wang2022probabilistic} & 0.305 & 0.409 \\
        \midrule
        \rowcolor{gray!15}\textbf{AgentWorld} & \textbf{0.318} & \textbf{0.416} \\
        \bottomrule
        \end{tabular}
        % \vspace{-10pt}
    \end{minipage}
    \hfill
    \begin{minipage}[t]{0.53\linewidth}
        % \vspace{-5pt}
        \centering
        \makeatletter\def\@captype{table}
        \setlength{\tabcolsep}{1.0mm}
        \caption{Action prediction performance on the nuScenes dataset measured by F1 score (\%). ${\text{\textdagger}}$ denotes models trained on nuScenes.}
        \label{tab:action}
        % \vspace{3pt}
        \begin{tabular}{l|ccc|cccc}
        \toprule
        \multirow{2}{*}{\textbf{Method}} 
        & \multicolumn{3}{c|}{\textbf{Lateral (F1)} $\uparrow$} 
        & \multicolumn{4}{c}{\textbf{Longitudinal (F1)} $\uparrow$} \\
        & forward & left & right & keep & acc. & dec. & stop \\
        \midrule
        Qwen2-VL-2B  & 62.43 & 22.75 & 28.65 & 40.70 & 50.23 & 49.21 & 41.04 \\
        Qwen2-VL-2B${\text{\textdagger}}$ & 92.60 & 61.78 & 66.52 & 56.42 & 74.32 & 76.10 & 74.85 \\
        VLA-World & \textbf{95.88} & 74.22 & 75.06 & 60.98 & 81.42 & 80.04 & 81.24 \\
        \midrule
        \rowcolor{gray!15}\textbf{AgentWorld} & 95.14 & \textbf{77.63} & \textbf{78.41} & \textbf{63.85} & \textbf{84.17} & \textbf{82.93} & \textbf{84.56} \\
        \bottomrule
        \end{tabular}
        % \vspace{-10pt}
    \end{minipage}
\end{table}
\paragraph{Open-loop planning.}
Tab.~\ref{tab:plan} shows that AgentWorld achieves highly competitive results without additional ego-state input, outperforming most prior VLA and VLA world baselines in both L2 error and collision rate.
When ego-status is further incorporated, AgentWorld obtains the best overall performance, reaching the lowest average L2 error and collision rate under both evaluation protocols.
Compared with VLA-World, the improvement is especially clear in long-horizon prediction and safety-critical collision metrics, suggesting that the agent-centric world state provides more reliable cues for future motion reasoning.
These results demonstrate that explicitly modeling surrounding agents yields stronger and safer planning than relying on future imagination or generic VLA reasoning alone. The qualitative results are shown in Fig.~\ref{fig:vis_com}. 
The visualized trajectories further show that AgentWorld produces more reliable waypoints in complex scenes with fewer tokens.

\paragraph{Perception results.}
As shown in Tab.~\ref{tab:perception}, AgentWorld achieves the best performance on mAP and NDS. This consistent improvement suggests that the proposed agent-centric latent states are not merely useful for downstream planning but also retain physically grounded 3D information required for perception. More importantly, AgentWorld improves detection while using agent tokens as the intermediate world state. This indicates that explicitly grounding each latent agent to physical participants provides a structured representation that benefits spatial understanding.

\paragraph{Meta-action results.} 
% \begin{wraptable}{r}{0.53\textwidth}
% \vspace{-4pt}
% \setlength{\tabcolsep}{3pt}
% \caption{Action prediction performance on the nuScenes dataset measured by F1 score (\%). ${\text{\textdagger}}$ denotes models trained on nuScenes.}
% \label{tab:action}
% \centering
% \resizebox{0.5\textwidth}{!}{
% \begin{tabular}{l|ccc|cccc}
%         \toprule
%         \multirow{2}{*}{Method} & \multicolumn{3}{c|}{Lateral $($F1$)$ $\uparrow$} & \multicolumn{4}{c}{Longitudinal $($F1$)$ $\uparrow$} \\
%         & forward & left & right & keep & acc. & dec. & stop \\
%         \midrule
%         Qwen2-VL-2B  & 62.43 & 22.75 & 28.65 & 40.70 & 50.23 & 49.21 & 41.04  \\
%         % \midrule
%         Qwen2-VL-2B${\text{\textdagger}}$ & 92.60 & 61.78 & 66.52 & 56.42 & 74.32 & 76.10 & 74.85 \\
%         % \midrule
%         VLA-World & 95.88 & 74.22 & 75.06 & 60.98 & 81.42 & 80.04 & 81.24 \\
%         \midrule
%         \rowcolor{gray!15}\textbf{AgentWorld} & 95.14 & 77.63 & 78.41 & 63.85 & 84.17 & 82.93 & 84.56 \\
%         \bottomrule
%         \end{tabular}
% }
% \end{wraptable}
Tab.~\ref{tab:action} 
reports action prediction performance on nuScenes.
AgentWorld achieves the best overall results across most lateral and longitudinal action categories.
The gains are especially pronounced for turning, lane keeping, acceleration, deceleration, and stopping, indicating that agent-centric world states provide more informative cues for both directional and speed-control decisions.
Although VLA-World obtains a slightly higher score on the relatively saturated \emph{forward} category, AgentWorld delivers more balanced improvements across interaction-sensitive actions.
These results show that explicitly grounding and reasoning over surrounding agents leads to stronger and more reliable policy prediction.

\paragraph{Closed-loop planning.}
Tab.~\ref{tab:bench2drive} reports closed-loop results on Bench2Drive.
AgentWorld achieves the highest driving score and success rate among all compared methods, demonstrating stronger robustness in interactive simulation.
It also obtains the best comfort score, suggesting that agent-centric reasoning improves not only task completion but also trajectory smoothness.
Although VAD has a slightly higher efficiency score, AgentWorld provides a better overall balance across safety, success, and comfort, validating the effectiveness of agent-centric world states for closed-loop driving.
% The closed-loop evaluation on Bench2Drive demonstrates that AgentWorld achieves stronger overall driving performance than existing end-to-end and modular driving baselines. Compared with prior methods, AgentWorld shows consistent advantages in task completion and driving quality, indicating that its world modeling capability leads to more effective decision-making in interactive traffic scenarios. Notably, the improvement is not limited to a single metric: AgentWorld maintains strong performance across driving score, success rate, efficiency, and comfort, suggesting a better balance between goal-directed navigation and human-like driving behavior. These results highlight the importance of learning an agent-centric world representation for robust closed-loop autonomous driving.

\begin{table*}[t]
\centering
\caption{Closed-loop evaluation results on Bench2Drive (CARLA) with SOTA methods.}
% \vspace{-2pt}
\label{tab:bench2drive}
\setlength{\tabcolsep}{6pt}
\resizebox{0.85\textwidth}{!}{
\begin{tabular}{lcccc}
\toprule
\textbf{Method} 
& \textbf{Driving Score $\uparrow$} 
& \textbf{Success Rate (\%) $\uparrow$} 
& \textbf{Efficiency $\uparrow$} 
& \textbf{Comfortness $\uparrow$} \\
\midrule
AD-MLP~\cite{zhai2023rethinking}        & 18.05 & 0.00  & 48.45  & 22.63 \\
UniAD-Base~\cite{hu2023planning}   & 45.81 & 16.36 & 129.21 & 43.58 \\
VAD~\cite{jiang2023vad}          & 42.35 & 15.00 & \textbf{157.94} & {46.01} \\
TCP-traj~\cite{wu2022trajectory}     & 59.90 & 30.00 & 76.54  & 18.08 \\
DriveAdapter~\cite{jia2023driveadapter} & 64.22 & 33.08 & 70.22  & 16.01 \\
Orion~\cite{fu2025orion}        & 77.74 & 54.62 & 151.48 & 17.38 \\
AutoVLA~\cite{zhou2025autovla}      & 78.84 & 57.73 & 146.93 & 39.33 \\
\midrule
\rowcolor{gray!15}
\textbf{AgentWorld}              & \textbf{81.17} & \textbf{61.36} & 147.82 & \textbf{54.39}\\
\bottomrule
\end{tabular}
}
% \vspace{-5pt}
\end{table*}

\subsection{Ablation Study}

\paragraph{Multimodal routed anchoring.}
Tab.~\ref{tab:mra} ablates the grounding components in AgentWorld.
Removing either 4D geometry or 2D visual anchoring degrades planning performance, indicating that both geometric structure and visual evidence are necessary for reliable agent grounding.
The full model achieves the lowest L2 error, showing that multimodal routed anchoring provides a more accurate and physically grounded agentic world state for planning. 
We also found that removing the 4D geometry features still yielded results comparable to the previous SOTA method~\cite{wang2026learning}, which proves the effectiveness of the agent-centric world state.

\paragraph{Policy learning strategies.}
Tab.~\ref{tab:two_stage} ablates policy learning strategies of AgentWorld. Directly using Qwen2-VL-2B leads to poor planning performance, while SFT (AgentWorld w/o RL) substantially reduces the L2 error. Further applying RL (AgentWorld) improves performance, showing that agent-centric policy optimization helps align the learned world state with planning objectives. These results validate the necessity of both supervised alignment and RL-based refinement in AgentWorld.
% The ablation on policy learning strategies shows that directly using the base vision-language model is insufficient for accurate agent behavior prediction. Supervised fine-tuning substantially improves the policy by aligning the model with task-specific demonstrations, while the subsequent RL stage further refines the behavior through closed-loop feedback. This indicates that AgentWorld benefits from a progressive learning pipeline: SFT provides a strong initialization, and RL improves policy quality by optimizing for interaction-aware objectives beyond imitation.

\begin{table}[t]
    \scriptsize
    \centering
    % \vspace{-5pt}
    \begin{minipage}[t]{0.28\linewidth}
    % \vspace{0pt}
    \centering
    \makeatletter\def\@captype{table}
    \setlength{\tabcolsep}{1.5mm}{
        \caption{Ablation on multimodal routed anchoring.}
        \label{tab:mra}
        % \vspace{3pt}
        \begin{tabular}{lc}
        \toprule 
        \textbf{Method} & \textbf{Avg. L2 (m)} $\downarrow$ \\
        \midrule
        w/o 4D geometry & 0.30 \\
        w/o 2D visual   & 0.28 \\
        \rowcolor{gray!15}AgentWorld & 0.27 \\
        \bottomrule
        \end{tabular}
        % \vspace{-5pt}
    }
    \end{minipage}%
    \hspace{10pt}%
    \begin{minipage}[t]{0.32\linewidth}
    % \vspace{0pt}
    \centering
    \makeatletter\def\@captype{table}
    \setlength{\tabcolsep}{1.2mm}{
        \caption{Ablation on different learning strategies.}
        \label{tab:two_stage}
        % \vspace{3pt}
        \begin{tabular}{lc}
        \toprule 
        \textbf{Method} & \textbf{Avg. L2 (m)} $\downarrow$ \\
        \midrule
        Qwen2-VL-2B       & 1.37 \\
        Qwen2-VL-2B + SFT & 0.32 \\
        \rowcolor{gray!15}Qwen2-VL-2B + SFT + RL & 0.27 \\
        \bottomrule
        \end{tabular}
        % \vspace{-5pt}
    }
    \end{minipage}%
    \hspace{10pt}%
    \begin{minipage}[t]{0.33\linewidth}
    % \vspace{0pt}
    \centering
    \makeatletter\def\@captype{table}
    \setlength{\tabcolsep}{1.0mm}{
        \caption{Latency of different VLA world models w/o ego.}
        \label{tab:latency}
        % \vspace{3pt}
        \begin{tabular}{lcc}
        \toprule 
        \textbf{Method} & \textbf{Avg. L2 (m)} $\downarrow$ & \textbf{Latency (s)} $\downarrow$\\
        \midrule
        FSDrive    & 0.53 & 0.36 \\
        VLA-World  & 0.30 & 0.40 \\
        \rowcolor{gray!15}AgentWorld & 0.27 & 0.31 \\
        \bottomrule
        \end{tabular}
        % \vspace{-5pt}
    }
    % \vspace{-25pt}
    \end{minipage}
\end{table}

\paragraph{Efficiency of VLA world models.}
% Tab.~\ref{tab:latency} compares the efficiency of different VLA world models.
% AgentWorld achieves the lowest L2 error while also having the shortest latency, outperforming both FSDrive and VLA-World.
% This shows that reasoning over compact agent-centric world states is not only more accurate but also more efficient than image-level future prediction.
% The result highlights the advantage of AgentWorld in balancing planning performance and inference efficiency.
Tab.~\ref{tab:latency} compares the efficiency of different VLA world models.
AgentWorld achieves the lowest L2 error while also having the shortest latency, outperforming both FSDrive and VLA-World.
This shows that reasoning over compact agent-centric world states is not only more accurate but also more efficient than image-level future prediction.
The result highlights the advantage of AgentWorld in balancing planning performance and inference efficiency, with potential benefits for practical autonomous driving.

% The ablation on model variants shows that AgentWorld is not simply determined by model scale, but also depends on the compatibility between the vision-language backbone and the downstream agent-world modeling objective. Different Qwen-VL variants achieve competitive results, suggesting that the proposed framework is broadly transferable across backbone choices. This robustness indicates that the performance gains mainly come from the AgentWorld training and reasoning formulation, rather than from a specific pretrained model alone.

% \begin{wrapfigure}{r}{0.42\textwidth}
% \vspace{-15pt}
% \centering
%     \includegraphics[width=1.0\linewidth]{figures/agent_num.pdf}
%     \vspace{-15pt}
%     \caption{Ablations on different agent numbers.}
%     \label{fig:agent_num}
% \end{wrapfigure}

\begin{figure}[t]
    \centering
    \includegraphics[width=1.0\linewidth]{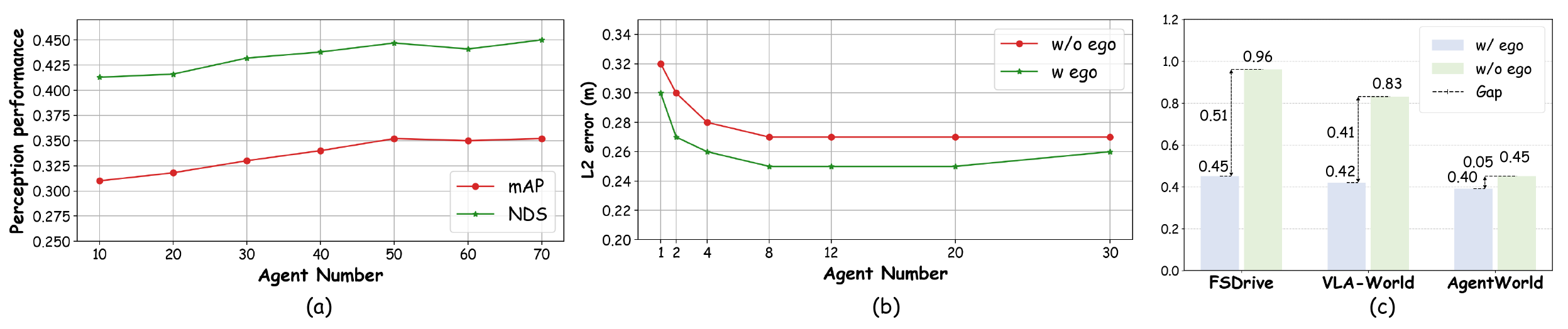}
    % \vspace{-12pt}
    \caption{Ablation studies on agent-centric world states. (a) Perception and (b) planning performance with varying numbers of agent tokens. (c) Performance comparisons of w/ and w/o ego-status.}
    \label{fig:agent}
    % \vspace{-15pt}
\end{figure}

\paragraph{Agent-centric world state.}
% Fig.~\ref{fig:agent} validates the effectiveness of agent-centric world states.
% Increasing the number of agent tokens consistently improves perception and reduces planning errors before saturation, indicating that AgentWorld benefits from modeling key surrounding agents.
% Moreover, removing ego-status causes only a small degradation compared to prior VLA world models, suggesting reduced reliance on ego-status shortcuts and stronger reasoning over grounded agent interactions.
Fig.~\ref{fig:agent} validates the effectiveness of agent-centric world states. Increasing the number of agent tokens consistently improves perception and reduces planning errors before saturation, indicating that AgentWorld benefits from modeling key surrounding agents. Moreover, removing ego-status causes only a small degradation compared to prior VLA world models, suggesting reduced reliance on ego-status shortcuts and stronger reasoning over grounded agent interactions when predicting the ego vehicle's future trajectory.

% The reward ablation confirms that each component in the RL objective contributes to the final policy. Removing trajectory-level, action-level, or agent-level rewards leads to degraded performance, showing that effective policy optimization requires supervision from multiple behavioral perspectives. In particular, trajectory and action rewards provide direct guidance for motion accuracy and decision consistency, while the agent-level reward encourages more globally coherent behavior. Together, these rewards form a complementary objective that stabilizes RL training and improves agent-centric prediction quality.

\begin{table}[t]
    \centering
    \begin{minipage}[t]{0.46\linewidth}
        \centering
        \small
        \caption{
            Representation comparisons on nuScenes with the same backbone, agent information, supervision, and SFT+RL training.
        }
        \label{tab:representation_controls}

        \setlength{\tabcolsep}{5pt}
        \begin{tabular}{lcc}
            \toprule
            Representation
            & L2 (m)$\downarrow$
            & Col. (\%)$\downarrow$ \\
            \midrule
            Global latent
            & 0.49 & 0.23 \\
            Structured text
            & 0.58 & 0.21 \\
            Agent tokens
            & \textbf{0.27} & \textbf{0.08} \\
            \bottomrule
        \end{tabular}
    \end{minipage}
    \hfill
    \begin{minipage}[t]{0.5\linewidth}
        \centering
        \small
        \caption{
            Agent-token interventions on nuScenes. Consistent permutations preserve token–agent correspondence; mismatched permutations disrupt it.
        }
        \label{tab:token_interventions}

        \setlength{\tabcolsep}{3pt}
        \begin{tabular}{lcc}
            \toprule
            Intervention
            & L2 (m)$\downarrow$
            & Col. (\%)$\downarrow$ \\
            \midrule
            None
            & 0.27 & 0.08 \\
            Consistent perm.
            & $0.27 \pm 0.01$
            & $0.08 \pm 0.03$ \\
            Mismatched perm.
            & $0.63 \pm 0.04$
            & $0.74 \pm 0.09$ \\
            \bottomrule
        \end{tabular}
    \end{minipage}
\end{table}

\subsection{Further Discussion}

\paragraph{Controlled representation comparisons.}
Tab.~\ref{tab:representation_controls} compares three representations under the same backbone, agent information, supervision, and SFT+RL training. Agent tokens achieve the lowest planning L2 error and collision rate, substantially outperforming both global latents and structured text. These gains across both metrics suggest that organizing world states around individual agents supports more accurate trajectory prediction and safer planning under the same training conditions. Structured text yields fewer collisions than global latents but incurs higher L2 error, whereas agent tokens improve both metrics. These results suggest that explicitly organizing world states around individual agents provides a more effective representation for planning.

\paragraph{Agent-token interventions.}
Tab.~\ref{tab:token_interventions} examines whether planning relies on the correspondence between agent tokens and their associated physical states. Consistent permutations preserve this correspondence and leave planning performance largely unchanged, whereas mismatched permutations substantially increase both trajectory error and collision rate. This contrast suggests that the planner relies on grounded token--agent associations rather than arbitrary slot indices.
The pronounced rise in collisions further suggests that preserving
these associations is important for reasoning about interactions
with surrounding agents and producing safe trajectories.

% \subsection{Visualization}
% \section{Conclusion}
% We presented AgentWorld, an agent-centric VLA world model for end-to-end autonomous driving. Our central claim is that the intermediate world state is a fundamental bottleneck in current VLA world models: without a physically grounded and interaction-aware state, strong imitation performance can still arise from shortcut signals rather than causal traffic dynamics. AgentWorld addresses this issue by representing the scene as a set of discrete agent-level latent tokens, each corresponding to a traffic participant that evolves over time in a 4D causal world state.
% By combining agent-level geometric grounding, multimodal feature integration, and agentic reinforcement learning, AgentWorld provides a structured interface between perception, world modeling, and decision reasoning. Extensive open-loop and closed-loop results demonstrate improved planning performance, stronger robustness to shortcut-prone settings, and more interpretable world states. These findings highlight that future VLA world models should not only imagine the future, but also define the right latent state over which driving policies can reason.
\section{Conclusion}
\label{sec:conlusion}

We presented \textbf{AgentWorld}, an agent-centric VLA world model for end-to-end autonomous driving. 
Instead of relying on visually plausible future imagination alone, AgentWorld represents the driving scene through discrete agent tokens that are dynamically grounded to surrounding participants. 
These tokens encode semantic identity, 3D geometry, motion, and interaction, providing a compact and planning-oriented latent world state beyond images. 
Through explicit geometric grounding, reasoning-oriented supervised fine-tuning, and agent-oriented reinforcement learning, AgentWorld learns to reason over agents and output future trajectories. 
Extensive open-loop and closed-loop experiments demonstrate that this agent-centric formulation improves planning performance, reduces shortcut reliance on ego-status extrapolation and language priors, and enables more interpretable world states. 
Despite these promising results, AgentWorld still relies on explicit geometric supervision to ground $\langle \mathrm{agent}\_i\rangle$ tokens to physical participants, rather than allowing agent tokens to emerge purely from end-to-end planning supervision.
Future work will explore more scalable self-supervised grounding with stronger visual-semantic binding, aiming to maintain consistent agent identities, physical states, and interaction semantics over time.
We hope AgentWorld provides a step toward physically grounded and interaction-aware world modeling for autonomous driving.

\section*{Acknowledgements}
This work was supported by the National Natural Science Foundation of China (Grant Nos. 62322113, 62376156), as well as the Shanghai Municipal Special Program for Basic Research on General AI Foundation Models (Grant No. 2025SHZDZX025G15).
% We hope AgentWorld provides a step toward physically grounded and interaction-aware world modeling for autonomous driving.

% \section{Limitations}
% While AgentWorld demonstrates the effectiveness of agent-centric world states, its current grounding stage still relies on explicit geometric supervision to bind special tokens to physical traffic participants. This design provides stable and interpretable agent representations, but it is not yet the most minimal formulation in which $\langle \mathrm{agent}\_i\rangle$ tokens emerge purely from end-to-end planning supervision. Moreover, the present grounding mainly emphasizes 3D geometry and motion; extending it to stronger visual-semantic binding could further improve the consistency between token identities, physical states, and interaction semantics over time. We view these directions as natural extensions toward more scalable and self-supervised agentic world modeling.

\bibliographystyle{unsrtnat}
\bibliography{ref}

%%%%%%%%%%%%%%%%%%%%%%%%%%%%%%%%%%%%%%%%%%%%%%%%%%%%%%%%%%%%

\newpage
\appendix

\section{Appendix}

\subsection{Data Curation}

This part describes the data curation pipeline used to train AgentWorld. 
The pipeline is designed around three principles. 
First, physical agent states are used only as supervision or reward signals, and are never exposed as textual input to the policy. 
Second, the \texttt{<perception>} field only exposes special agent tokens, while the state of each participant is carried by the token hidden state. 
Third, the final SFT phase and the RL phase share the same MLLM-only input-output template.

\paragraph{Influential agent selection.}
Since AgentWorld uses a fixed number of agent tokens, we select the most planning-relevant participants for grounding and reasoning. 
For each sample, we rank annotated objects by an influence score that considers both their current state and future interaction with the ego vehicle. 
The selected agents are used as grounding targets in Stage 1 and as the source of agent-token references in Stage 2.

Let $\mathcal{T}_i=\{\mathbf{p}_i^{t+\Delta_m}\}_{m=1}^{M_H}$ denote the future trajectory of object $i$ within a horizon of $H=3$s, and let $\mathcal{T}_{\mathrm{ego}}=\{\mathbf{p}_{\mathrm{ego}}^{t+\Delta_m}\}_{m=1}^{M_H}$ denote the expert ego trajectory. 
We define the relative future trajectory as
\begin{equation}
    \mathbf{r}_i^{t+\Delta_m}
    =
    \mathbf{p}_i^{t+\Delta_m}
    -
    \mathbf{p}_{\mathrm{ego}}^{t+\Delta_m}.
\end{equation}
The current influence score measures the relevance of an object based on its current position, lane relation, distance, and closing speed:
\begin{equation}
    S_i^{\mathrm{cur}}
    =
    w_d s_i^{\mathrm{dist}}
    +
    w_l s_i^{\mathrm{lane}}
    +
    w_c s_i^{\mathrm{close}},
\end{equation}
where $s_i^{\mathrm{dist}}$ increases for nearby objects, $s_i^{\mathrm{lane}}$ indicates whether the object is in or close to the ego lane, and $s_i^{\mathrm{close}}$ measures relative closing speed.

The future influence score measures whether the object is likely to affect the ego plan:
\begin{equation}
    S_i^{\mathrm{future}}
    =
    w_{\mathrm{min}} s_i^{\mathrm{min}}
    +
    w_{\mathrm{ttc}} s_i^{\mathrm{ttc}}
    +
    w_{\mathrm{corr}} s_i^{\mathrm{corr}},
\end{equation}
where $s_i^{\mathrm{min}}$ is based on the minimum distance between $\mathcal{T}_i$ and $\mathcal{T}_{\mathrm{ego}}$, $s_i^{\mathrm{ttc}}$ measures time-to-collision risk, and $s_i^{\mathrm{corr}}$ indicates whether the future object trajectory enters the ego corridor.

The final influence score is
\begin{equation}
    S_i
    =
    \alpha_i
    \left(
    \lambda S_i^{\mathrm{cur}}
    +
     (1-\lambda)S_i^{\mathrm{future}}\right),
\end{equation}
where $\alpha_i$ is a category-dependent weight. $\lambda$ is set to 0.45.
We use larger weights for vulnerable road users such as pedestrians and cyclists, since they are safety-critical even when their current distance is not the smallest.

The algorithm is shown in Alg.~\ref{alg:topk_agent_selection}
In our implementation, $K\leq N$, where $N$ is the number of available agent tokens. 
If fewer than $K$ valid objects exist, the remaining slots are treated as no-object. 
This selection is used to prioritize agents that are most relevant to planning, rather than exhaustively supervising all annotated objects in crowded scenes.

\begin{figure}[t]
    \centering
    \includegraphics[width=1.0\linewidth]{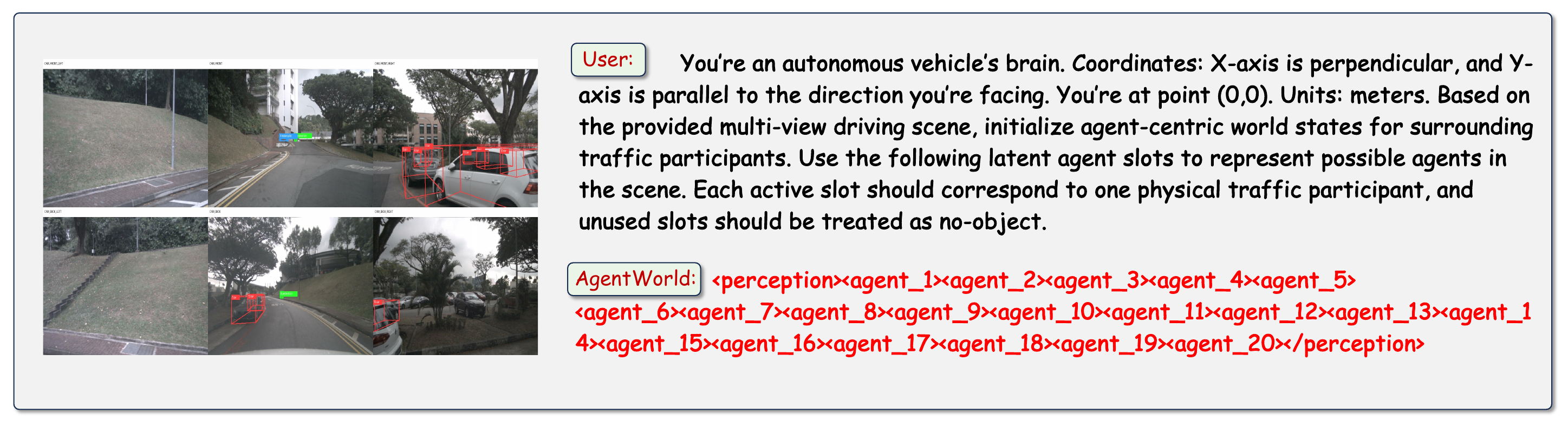}
    % \vspace{-10pt}
    \caption{Example of the Stage-1 agent grounding QA template. Given multi-view driving observations and a grounding instruction, AgentWorld outputs a fixed set of \texttt{<agent\_i>} tokens inside the \texttt{<perception>} field. These tokens act as latent agent slots: active slots are supervised to align with physical traffic participants through 3D box and future trajectory targets, while unused slots are trained as no-object.}
    \label{fig:grounding_data}
    % \vspace{-10pt}
\end{figure}

\paragraph{Stage 1: grounding data.}
The goal of Stage 1 is to turn the special tokens \(\{\langle\mathrm{agent}\_i\rangle\}_{i=1}^{N}\) into physically grounded world-state slots. 
For each scene, we construct a grounding sample with the following visible input shown in Fig.~\ref{fig:grounding_data}:
\begin{equation}
    \texttt{<perception>}
    \langle\mathrm{agent}\_1\rangle
    \cdots
    \langle\mathrm{agent}\_N\rangle
    \texttt{</perception>}.
\end{equation}
No textual description of object category, location, or motion is provided. 
The model must encode this information in the contextual hidden states of the agent tokens.

We compute influence scores using Alg.~\ref{alg:topk_agent_selection}. Only top-K agents are used.
The training targets are the selected influential agents \(\mathcal{A}_K\). 
For each selected agent, we store its category, 3D bounding box, velocity or motion state, and future trajectory. 
During training, the predictions from the agent-token hidden states are matched to these targets using DETR-style bipartite matching. 
Matched tokens are supervised with classification, 3D box, and trajectory losses, while unmatched tokens are supervised as no-object. 
The offline VGGT features are provided as geometry memory for routed anchoring, but the VGGT encoder itself is not trained. 

\begin{figure}[t]
    \centering
    \includegraphics[width=1.0\linewidth]{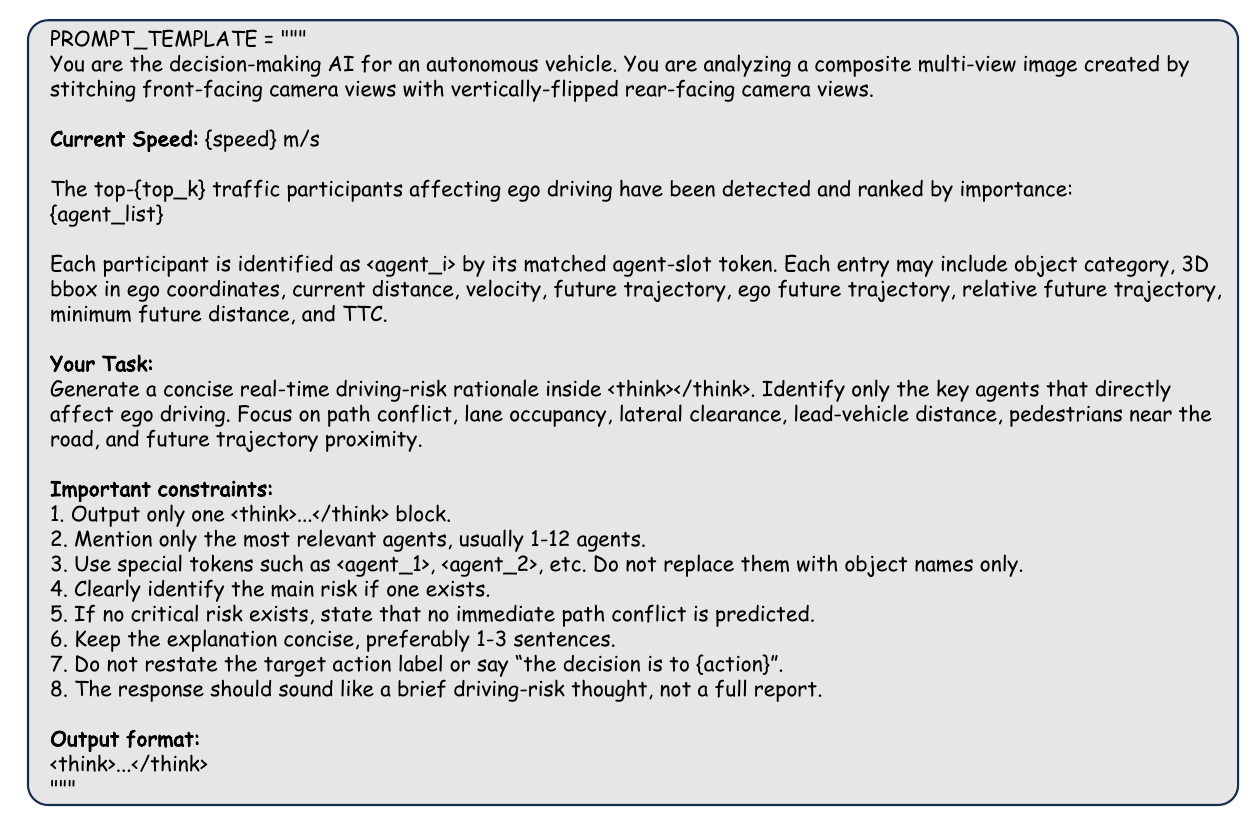}
    % \vspace{-15pt}
    \caption{Prompt design for reasoning with agent data generation.}
    \label{fig:vis_prompt}
    % \vspace{-20pt}
\end{figure}

\paragraph{Stage 2: Reasoning-oriented SFT data.}
Stage 2 teaches the MLLM to use grounded agent tokens for traffic reasoning and planning. 
Each response follows a compact structured format:
\begin{equation}
    y=(\xi,a,\tau),
\end{equation}
where \(\xi\) is a short rationale in \texttt{<think>}, \(a\) is a high-level action in \texttt{<action>}, and \(\tau\) is the future ego trajectory in \texttt{<answer>}.

\begin{figure}[t]
    \centering
    \includegraphics[width=1.0\linewidth]{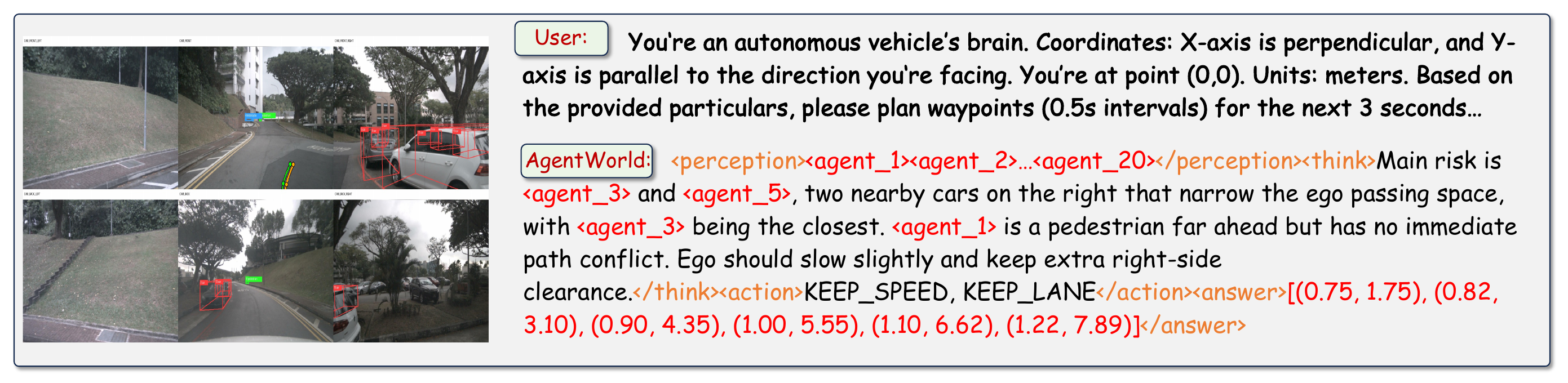}
    % \vspace{-10pt}
    \caption{Example of the Stage-2 reasoning-oriented SFT template. Given multi-view driving observations, ego-state instructions, and the fixed \texttt{<agent\_i>} slots in the \texttt{<perception>} field, AgentWorld learns to generate a structured response consisting of concise agent-grounded reasoning in \texttt{<think>}, a high-level driving command in \texttt{<action>}, and future ego waypoints in \texttt{<answer>}. The rationale explicitly references decision-relevant agent tokens, encouraging the policy to plan based on surrounding-agent interactions rather than shortcut cues.
}
    \label{fig:sft_data}
    % \vspace{-10pt}
\end{figure}

As shown in Fig.~\ref{fig:sft_data}, the SFT prompt uses the same \texttt{<perception>} field as Stage 1:
\begin{equation}
    \texttt{<perception>}
    \langle\mathrm{agent}\_1\rangle
    \cdots
    \langle\mathrm{agent}\_N\rangle
    \texttt{</perception>}.
\end{equation}

% \begin{equation}
% \small
% \begin{aligned}
% &\texttt{<think>}
% \text{ \texttt{<agent\_3>} may enter the ego path, so the ego vehicle should decelerate and keep lane.}
% \texttt{</think>} \\
% &\texttt{<action>DECELERATE, KEEP\_LANE</action>} \\
% &\texttt{<answer>}[(x_1,y_1),\ldots,(x_H,y_H)]\texttt{</answer>}.
% \end{aligned}
% \end{equation}

To generate the rationale, we use the slot-to-agent assignment from the Stage-1 grounding model. 
Specifically, we run the grounding model on the training set, match predicted slots to ground-truth agents, and obtain a mapping from \(\langle\mathrm{agent}\_i\rangle\) to either a physical agent or no-object. 
% We then compute influence scores using Alg.~\ref{alg:topk_agent_selection}. 
% Only high- and medium-risk tokens are referenced in \texttt{<think>}; low-risk or no-object tokens are either omitted or summarized as non-critical.

The high-level action is derived from the expert ego trajectory. 
We classify longitudinal behavior into \texttt{ACCELERATE}, \texttt{KEEP\_SPEED}, \texttt{DECELERATE}, or \texttt{STOP}, and lateral behavior into \texttt{KEEP\_LANE}, \texttt{CHANGE\_LEFT}, or \texttt{CHANGE\_RIGHT}. 
The \texttt{<answer>} field contains the expert future waypoints.

We optionally use strong MLLMs to generate concise, agent-grounded driving-risk rationales. The prompt template is shown in Fig.~\ref{fig:vis_prompt}. The annotator model receives multi-view driving images, the current ego speed, and structured information about relevant traffic participants identified by their matched agent-slot tokens, and generates a brief rationale in <think> focusing on the key agents and their implications for driving risk.

% Raw track IDs and exact future labels are not exposed. 

\paragraph{Stage 3: RL Data}

The RL stage uses the same prompt template as the final SFT phase, but removes the supervised assistant response. 
The model observes images, ego status, navigation command, and the \texttt{<perception>} field containing the agent tokens. 
It then samples multiple structured responses:
\begin{equation}
    y_k=(\xi_k,a_k,\tau_k),
    \quad k=1,\ldots,K_{\mathrm{RL}}.
\end{equation}

% The reward computation uses information that is not exposed to the model, including expert ego trajectory, future agent trajectories, map information, drivable area, traffic light state, and collision checking. 
We use a composite reward that measures format, perception, action, trajectory, and interaction context. 
This makes the RL stage agent-oriented: to improve reward, the model must generate trajectories that are safe with respect to the dynamic agents represented by the learned tokens.

% We oversample long-tail and interaction-heavy scenes during RL, including pedestrian crossing, cut-in, leading-vehicle braking, lane change, crowded intersection, and stop-and-go scenarios. 
% This prevents the model from exploiting dataset-level shortcuts such as always keeping lane or always maintaining speed.

\subsection{Rewards of RL}
\noindent\textbf{Format reward}.
This reward $R_{\mathrm{fmt}}$ enforces the hierarchical output protocol of AgentWorld. 
A structural reward of 1.0 is assigned if the response strictly follows
\texttt{<perception>}...\texttt{</perception>}
\texttt{<think>}...\texttt{</think>}
\texttt{<action>}...\texttt{</action>}
\texttt{<answer>}...\texttt{</answer>},
and 0 otherwise. 
The \texttt{<perception>} field must contain only valid special agent tokens 
$\{\langle\mathrm{agent}\_i\rangle\}_{i=1}^{N}$, without free-form object descriptions or invalid tokens. 
The \texttt{<action>} field must be parsable into a valid high-level maneuver, and the \texttt{<answer>} field must contain a complete sequence of future waypoints with the required horizon.  We also encourage the \texttt{<think>} tag to contain at least one agent token.
% The final format reward is the average of structural validity, agent-token validity, action validity, and trajectory completeness.

\noindent\textbf{Perception reward.}
AgentWorld does not require the policy to explicitly output bounding boxes or textual perception descriptions. 
Instead, perception is exposed through a fixed set of latent agent slots in the \texttt{<perception>} field. 
Therefore, we define a lightweight slot-alignment reward that checks whether the generated perception field contains the correct number of valid agent tokens. 
Given the predefined token set 
$\mathcal{T}_{\mathrm{agent}}=\{\langle\mathrm{agent}_1\rangle,\cdots,\langle\mathrm{agent}_N\rangle\}$ 
and the parsed token set $\hat{\mathcal{T}}_{\mathrm{agent}}$ from the response, the reward is defined as
\begin{equation}
    R_{\mathrm{percep}}
    =
    \mathbb{I}
    \left[
    |\hat{\mathcal{T}}_{\mathrm{agent}}| = N
    \right]
    \cdot
    \mathbb{I}
    \left[
    \hat{\mathcal{T}}_{\mathrm{agent}} = \mathcal{T}_{\mathrm{agent}}
    \right],
\end{equation}
where duplicate, missing, or invalid tokens receive zero reward. 
We additionally require the \texttt{<perception>} field to contain only agent tokens and no free-form object descriptions. 
This reward ensures that the policy preserves the agent-token interface learned during grounding, while leaving the physical content of each slot to be represented by its hidden state rather than by generated text.

\noindent\textbf{Meta-Action reward.}
We evaluate the high-level driving command predicted in the \texttt{<action>} field. 
The action is represented as a set of discrete maneuver labels, such as 
\texttt{KEEP\_LANE}, \texttt{CHANGE\_LEFT}, \texttt{DECELERATE}, or \texttt{STOP}. 
Given the reference action set $\mathcal{M}^{\star}$ derived from the expert trajectory and the predicted action set $\hat{\mathcal{M}}$, we compute an F1-based reward:
\[
    R_{\mathrm{action}}
    =
    \frac{
    2\cdot \mathrm{Prec}(\hat{\mathcal{M}},\mathcal{M}^{\star})
    \cdot \mathrm{Rec}(\hat{\mathcal{M}},\mathcal{M}^{\star})
    }{
    \mathrm{Prec}(\hat{\mathcal{M}},\mathcal{M}^{\star})
    +
    \mathrm{Rec}(\hat{\mathcal{M}},\mathcal{M}^{\star})
    +
    \epsilon
    } .
\]
This balances precision and recall over multi-label driving intents and prevents degenerate behaviors such as always predicting conservative or straight-driving actions.

\noindent\textbf{Agent interaction reward.}
We use the frozen agent head as a reward-side probe to decode the top-$K$ active agent trajectories from the learned \texttt{<agent\_i>} tokens. 
Given the predicted ego trajectory $\hat{\tau}_{\mathrm{ego}}$ and decoded agent trajectories $\{\hat{\tau}_i\}_{i\in\mathcal{I}_K}$, we compute the minimum future distance
\begin{equation}
    d_i^{\min}
    =
    \min_m
    \left\|
    \hat{\mathbf{p}}_{\mathrm{ego}}^{t+\Delta_m}
    -
    \hat{\mathbf{p}}_{i}^{t+\Delta_m}
    \right\|_2 ,
\end{equation}
and the time-to-collision risk
\begin{equation}
    \mathrm{TTC}_i
    =
    \min
    \left\{
    \Delta_m
    \mid
    \left\|
    \hat{\mathbf{p}}_{\mathrm{ego}}^{t+\Delta_m}
    -
    \hat{\mathbf{p}}_{i}^{t+\Delta_m}
    \right\|_2
    <
    d_{\mathrm{safe}}
    \right\}.
\end{equation}
If no unsafe distance occurs within the horizon, we set $\mathrm{TTC}_i=\infty$. 
The agent reward penalizes collision, insufficient clearance, and short TTC:
\begin{equation}
    R_{\mathrm{agent}}
    =
    -
    \frac{1}{K}
    \sum_{i\in\mathcal{I}_K}
    \left(
    \lambda_{\mathrm{col}}\mathbb{I}[d_i^{\min}<d_{\mathrm{col}}]
    +
    \lambda_{\mathrm{clear}}
    \left[1-\frac{d_i^{\min}}{d_{\mathrm{safe}}}\right]_+
    +
    \lambda_{\mathrm{ttc}}
    \left[1-\frac{\mathrm{TTC}_i}{T_{\mathrm{safe}}}\right]_+
    \right).
\end{equation}
% The core code can be found in Alg.~\ref{alg:agent_reward_code}

\noindent\textbf{Overall reward.}
The final reward is a weighted sum:
\begin{equation}
    R
    =
    \lambda_{\mathrm{fmt}} R_{\mathrm{fmt}}
    +
    \lambda_{\mathrm{percep}} R_{\mathrm{percep}}
    +
    \lambda_{\mathrm{action}} R_{\mathrm{action}}
    +
    \lambda_{\mathrm{traj}} R_{\mathrm{traj}}
    +
    \lambda_{\mathrm{agent}} R_{\mathrm{agent}} .
\end{equation}
$R_{\mathrm{fmt}}$ and $R_{\mathrm{percep}}$ enforce valid structured generation and correct use of the fixed \texttt{<agent\_i>} slots. 
$R_{\mathrm{action}}$ and $R_{\mathrm{traj}}$ evaluate maneuver correctness and waypoint quality. 
$R_{\mathrm{agent}}$ further makes the reward interaction-aware by penalizing collisions, unsafe clearance, and short TTC between the predicted ego trajectory and the top-$K$ agent trajectories decoded from the learned agent tokens by a frozen agent head. 
Thus, the reward encourages AgentWorld to preserve its agent-token reasoning interface while optimizing safe driving behavior.

\subsection{Training Details}
Our training follows three stages: agent grounding SFT, reasoning-oriented SFT, and agent-oriented reinforcement learning. 
% All stages are trained with bfloat16 mixed precision. 
The offline VGGT features are pre-extracted.

\noindent\textbf{Stage 1: Physical grounding SFT.}
We first train the agent-token grounding interface for 12 epochs. 
The trainable modules include the MLLM backbone, the special \texttt{<agent\_i>} token embeddings, the multimodal routed anchoring module, and the auxiliary agent head. 
We use AdamW with a cosine decay schedule and a warmup ratio of 0.03. The learning rate is set to $1.0\times10^{-4}$. 
The per-device batch size is 2 with 8 gradient accumulation steps, giving an effective batch size of 16 per device. 
Input images are processed with a maximum resolution of 524288 pixels. For multimodal routed anchoring, we use $L=3$ routing layers by default. 
The grounding loss supervises category/existence, 3D bounding box, and future trajectory prediction after Hungarian matching. The number of agents is set to 20.

\noindent\textbf{Stage 2: Reasoning-oriented SFT.}
Starting from the Stage-1 checkpoint, we train the MLLM to use grounded agent tokens for reasoning and planning for 12 epochs. 
The multimodal routed anchoring module and auxiliary agent head are frozen. 
The trainable parameters are the MLLM policy parameters and the special agent-token embeddings. 
We use AdamW with a learning rate of $5.0\times10^{-5}$ and a cosine decay schedule. 
The batch configuration follows Stage 1, with a per-device batch size of 2 and 8 gradient accumulation steps. 
Each sample is trained with the structured output format consisting of \texttt{<perception>}, \texttt{<think>}, \texttt{<action>}, and \texttt{<answer>}. 
To reduce mismatch with the RL stage, we progressively drop the training-time grounding scaffold during SFT: early batches use full grounding inputs, middle batches use partial dropout, and the final SFT phase uses the same MLLM-only format as RL.

\noindent\textbf{Stage 3: Agent-oriented RL.}We further optimize the Stage-2 checkpoint with GRPO for 1 epoch. During RL, only the MLLM policy parameters are updated. The auxiliary agent head is kept frozen and used only as a reward-side generator to decode top-$K$ agent trajectories from the learned \texttt{<agent\_i>} tokens; it receives no gradients. The routed anchoring modules are also frozen and are not optimized during RL. We use a learning rate of $1.0\times10^{-6}$ and a global batch size of 16. For each prompt, we sample 8 responses to estimate group-relative advantages. The KL coefficient to the SFT reference policy is set to $1.0\times10^{-2}$. 
The reward is a weighted sum of format reward, perception-slot reward, action reward, trajectory reward, and agent interaction reward, with all reward weights set to 1 unless otherwise specified. 

After grounding and SFT, the agent tokens have already been aligned with physical participants and learned as the planning interface of the MLLM. 
Therefore, in the RL stage, we no longer optimize the auxiliary grounding modules. 
Starting from the SFT checkpoint, GRPO updates only the MLLM policy, which samples multiple structured responses for each scene, including a rationale, a high-level action, and future waypoints.
The auxiliary agent head is kept frozen only as a reward-side decoder. 
It decodes the hidden states of agent tokens into physical quantities, such as agent existence, 3D boxes, and short-horizon motions, which are then used to compute candidate-dependent interaction rewards for the sampled ego trajectories. 
This frozen decoder receives no gradient and is not part of the policy forward path. 
At inference time, it is completely removed, and AgentWorld performs planning with the MLLM and the learned agent-token interface alone.

% \paragraph{Training objectives.}
\paragraph{Training objectives.}
AgentWorld is optimized in three stages. 
First, we perform agent grounding by supervising the hidden state $\mathbf{h}_t^i$ at each agent-token position with a lightweight head that predicts category/existence, 3D state, and future trajectory. 
Because the token order is unordered, predicted slots are assigned to ground-truth agents with DETR-style bipartite matching. 
The grounding loss is
\begin{equation}
    \mathcal{L}_{\mathrm{ground}}
    =
    \lambda_{\mathrm{cls}}\mathcal{L}_{\mathrm{cls}}
    +
    \lambda_{\mathrm{box}}\mathcal{L}_{\mathrm{box}}
    +
    \lambda_{\mathrm{traj}}\mathcal{L}_{\mathrm{traj}}
    +
    \lambda_{\mathrm{noobj}}\mathcal{L}_{\mathrm{noobj}}.
\end{equation}
We set $\lambda_{\mathrm{cls}}=1.0$, $\lambda_{\mathrm{box}}=1.0$, $\lambda_{\mathrm{traj}}=1.0$, and $\lambda_{\mathrm{noobj}}=0.1$ by default. 
The no-object weight is kept small to avoid dominating the supervision when most agent slots are empty. 
Matched slots are grounded to physical agents, while unmatched slots are trained as no-object.

Second, we conduct reasoning-oriented SFT. 
Each expert response consists of a concise rationale, a high-level action, and future waypoints, denoted as $y^\star=(\xi^\star,a^\star,\tau^\star)$. 
The policy is trained with the autoregressive objective
\begin{equation}
    \mathcal{L}_{\mathrm{SFT}}
    =
    -
    \sum_k
    \log p_\theta
    \left(
    y_k^\star
    \mid
    y_{<k}^\star,
    \mathcal{H}_t,
    \mathcal{S}^{\mathrm{agent}}_t
    \right),
    \quad
    \mathcal{S}^{\mathrm{agent}}_t=\{\mathbf{h}_t^i\}_{i=1}^{N}.
\end{equation}
This stage teaches the MLLM to use grounded agent tokens for traffic saliency analysis, causal reasoning, and trajectory planning. 
When auxiliary grounding modules are retained in the early SFT phase, we optionally add a small grounding regularizer:
\begin{equation}
    \mathcal{L}_{\mathrm{Stage2}}
    =
    \mathcal{L}_{\mathrm{SFT}}
    +
    \eta \mathcal{L}_{\mathrm{ground}},
\end{equation}
where $\eta$ is linearly decayed from $0.1$ to $0$ as the training gradually shifts to the MLLM-only format. 
In the final SFT phase, $\eta=0$. Finally, we remove the training-time grounding scaffold and optimize the MLLM-only policy with GRPO.

\subsection{Comparison with Different Latent VLA World Models}
Recent latent VLA world models reduce the cost of future imagination by moving from pixel-space generation to latent-space reasoning. 
For example, Latent-CoT-style driving models~\cite{tan2025latent} replace verbose textual reasoning with latent world-model tokens, DriveWorld-VLA~\cite{liu2026driveworld} unifies world modeling and planning in a shared latent space, and LaST-VLA~\cite{luo2026last} performs reasoning in a latent spatio-temporal space with geometric and dynamic alignment. 
These methods improve efficiency over explicit future-frame generation, but their world states remain largely image-, BEV-, occupancy-, or spatio-temporal-latent representations. 
Such latents can still preserve fine-grained scene factors that are weakly related to planning, including appearance, background layout, static context, dense occupancy patterns, and generic scene evolution, while leaving the agent-level causal variables behind driving decisions implicit. 
In contrast, AgentWorld changes the unit of world modeling from dense scene evolution to agent evolution. 
Rather than compressing the future into a holistic latent state, AgentWorld represents the driving world as a set of grounded and addressable \texttt{<agent\_i>} tokens, where each token is supervised to correspond to a surrounding traffic participant or a no-object state and encodes semantic identity, 3D geometry, motion tendency, and interaction context. 
This agent-centric world state filters out much of the planning-irrelevant nuisance information retained by dense latent world models and exposes the physical entities and interactions that directly govern ego planning.

\subsection{Closed-Loop Evaluation Metrics}

We evaluate closed-loop driving performance following the standard protocol in Bench2Drive~\cite{jia2024bench2drive}. Unlike open-loop trajectory evaluation, closed-loop simulation measures how sequential decisions affect future observations
and safety outcomes. We report four metrics: success rate, driving score, efficiency, and comfortness.

\paragraph{Success Rate (SR).}
Success rate measures the fraction of episodes in which the ego vehicle reaches the target without safety-critical infractions:
\begin{equation}
    \mathrm{SR}
    =
    \frac{1}{N}
    \sum_{i=1}^{N}
    \mathbb{I}(S_i),
\end{equation}
where $N$ is the number of evaluated episodes and $\mathbb{I}(S_i)$ indicates whether episode $i$ is successfully completed.

\paragraph{Driving Score (DS).}
Driving score combines route completion and safety penalties:
\begin{equation}
    \mathrm{DS}
    =
    \frac{1}{N}
    \sum_{i=1}^{N}
    R_i \prod_k P_{i,k},
\end{equation}
where $R_i \in [0,1]$ is the route completion ratio, and $P_{i,k} \in (0,1]$ is the penalty multiplier for the $k$-th infraction, such as collision, red-light violation,
or off-road driving.

\paragraph{Efficiency.}
Efficiency measures the ego vehicle's speed relative to surrounding traffic at route checkpoints. We use the official Bench2Drive~\cite{jia2024bench2drive}  implementation to aggregate the speed percentages across checkpoints and routes. Higher values indicate greater
efficiency, and values above 100 indicate speeds exceeding those of the reference traffic.
% Efficiency evaluates whether the agent completes successful routes without unnecessary delay:
% \begin{equation}
%     \mathrm{Eff}
%     =
%     \frac{1}{|S|}
%     \sum_{i\in S}
%     \frac{T_i^{\mathrm{expert}}}{T_i^{\mathrm{agent}}},
% \end{equation}
% where $S$ is the set of successful episodes, $T_i^{\mathrm{agent}}$ is the traversal time of the evaluated agent, and $T_i^{\mathrm{expert}}$ is the expert or reference traversal time.

\paragraph{Comfortness.}
Comfortness measures the proportion of trajectory segments satisfying the benchmark's kinematic constraints on acceleration, jerk, yaw rate, and yaw acceleration. We use the official Bench2Drive~\cite{jia2024bench2drive} implementation and report the resulting score as a percentage. Higher values indicate smoother driving.

% \subsection{More Method Details}
\subsection{The Theoretical Motivation of AgentWorld}
We provide a theoretical perspective on why agent-centric latent states are more suitable than pixel-level future frames for planning-oriented VLA world models.
Let $\mathcal{H}_t$ denote the driving context and let $\tau$ be a future ego trajectory.
A pixel-level world model predicts future visual observations $\hat{\mathcal{I}}_{t+1:t+H}$, whereas AgentWorld predicts an agent-centric latent state $\hat{\mathcal{A}}_{t+1:t+H}$.
Each agent state encodes semantic identity, 3D geometry, motion state, and interaction context.

We view future images as a composition of two parts:
\begin{equation}
    \mathcal{I}_{t+1:t+H}
    =
    g(\mathcal{A}_{t+1:t+H}, \mathcal{U}_{t+1:t+H}),
\end{equation}
where $\mathcal{A}$ denotes the physical and interactive evolution of traffic participants, and $\mathcal{U}$ denotes nuisance visual factors such as texture, illumination, background, and visual continuity.
For planning, the cost of an ego trajectory is determined by the evolution of traffic agents rather than by all pixel-level details.
We therefore assume that
\begin{equation}
    C(\tau,\mathcal{S}) = C(\tau,\mathcal{A}),
\end{equation}
where $\mathcal{S}$ is the full future scene state and $\mathcal{A}$ is the agent-centric latent state.

Under this assumption, $\mathcal{A}$ serves as a sufficient statistic for planning.
For any policy conditioned on future images, we can construct an agent-conditioned policy by marginalizing out nuisance visual factors:
\begin{equation}
    \pi_{\mathcal{A}}(\tau \mid \mathcal{H}_t,\mathcal{A})
    =
    \mathbb{E}_{\mathcal{U}\mid\mathcal{A}}
    \left[
    \pi_{\mathcal{I}}(\tau \mid \mathcal{H}_t, g(\mathcal{A},\mathcal{U}))
    \right].
\end{equation}
Since the planning cost depends on $\mathcal{A}$ but not on $\mathcal{U}$, conditioning the policy on nuisance visual details does not provide additional decision-relevant information.
This does not imply that images are unnecessary; visual evidence remains essential for grounding and estimating $\mathcal{A}$.
Rather, it shows that once the agent-centric state is available, reasoning over dense generated pixels is an indirect and inefficient interface for planning.

We further show that planning performance is directly tied to the quality of the predicted agent-centric state.
Assume the planning cost is $L$-Lipschitz with respect to the agent-centric state:
\begin{equation}
    |C(\tau,\mathcal{A}) - C(\tau,\hat{\mathcal{A}})|
    \leq
    L d(\mathcal{A},\hat{\mathcal{A}}),
\end{equation}
where $d(\cdot,\cdot)$ measures the prediction error of the agent-centric state.
Let
\begin{equation}
    \hat{\tau}
    =
    \arg\min_{\tau} C(\tau,\hat{\mathcal{A}}),
    \quad
    \tau^\star
    =
    \arg\min_{\tau} C(\tau,\mathcal{A}).
\end{equation}
Then the regret of planning with the predicted agent-centric state is bounded by
\begin{equation}
    C(\hat{\tau},\mathcal{A}) - C(\tau^\star,\mathcal{A})
    \leq
    2L d(\mathcal{A},\hat{\mathcal{A}}).
\end{equation}
This bound shows that reducing the agent-state prediction error directly reduces planning regret.
In contrast, a low pixel reconstruction loss does not necessarily imply a low agent-state error, because the pixel objective also allocates representation capacity to nuisance factors $\mathcal{U}$.
AgentWorld therefore provides a more task-aligned surrogate for planning by explicitly grounding latent tokens to 3D geometry, motion dynamics, and interactions.

The computational benefit also follows from the compactness of the state.
If each future frame contains $P$ visual tokens, a pixel-level world model with horizon $H$ and $K$ iterative generation steps processes $O(KHP)$ visual tokens.
AgentWorld instead uses $N$ agent tokens per step, resulting in $O(HN)$ world-state tokens, where $N \ll P$.
When self-attention over generated world tokens dominates the computation, the complexity scales as
\begin{equation}
    O\big(K(HP)^2\big)
    \quad \text{versus} \quad
    O\big((HN)^2\big),
\end{equation}
leading to an approximate ratio of
\begin{equation}
    O\left(K\left(\frac{P}{N}\right)^2\right).
\end{equation}
Thus, AgentWorld concentrates computation on decision-relevant traffic participants rather than dense visual details, yielding a compact, structured, and planning-oriented world state for VLA-based autonomous driving.

\begin{algorithm}[t]
\caption{Top-$K$ Influential Agent Selection}
\label{alg:topk_agent_selection}
\begin{algorithmic}[1]
\Require Current sample annotations, ego pose, future horizon $H=3$s, number of selected agents $K$
\Ensure Top-$K$ influential agents $\mathcal{A}_K$
\For{each object $i$ in the current sample}
    \State Transform the object box, velocity, and category into the current ego frame.
    \State Extract the future object trajectory $\mathcal{T}_i$ from the annotation chain.
    \State Extract the future ego trajectory $\mathcal{T}_{\mathrm{ego}}$.
    \State Compute the relative future trajectory $\mathbf{r}_i^{t+\Delta_m}$.
    \State Compute current influence score $S_i^{\mathrm{cur}}$.
    \State Compute future trajectory influence score $S_i^{\mathrm{future}}$.
    \State Compute final score $S_i=\alpha_i(\lambda S_i^{\mathrm{cur}}+(1-\lambda)S_i^{\mathrm{future}})$.
\EndFor
\State Sort all objects by $S_i$ in descending order.
\State Select the top-$K$ objects as $\mathcal{A}_K$.
\State \Return $\mathcal{A}_K$
\end{algorithmic}
\end{algorithm}

\begin{table}[t]
    \scriptsize
    \scalebox{1}{
        \begin{minipage}[b]{0.33\linewidth}
        \centering
        \makeatletter\def\@captype{table}
        \setlength{\tabcolsep}{3mm}{
            \caption{Ablation on agent head in SFT and RL stages.}
            \label{tab:head}
            \vspace{3pt}
            \begin{tabular}{lc}
        \toprule 
        \textbf{Method} & \textbf{Avg. L2 (m)} $\downarrow$ \\
        \midrule
        SFT w/o head &  0.31\\
        RL w head &  0.26\\
        \rowcolor{gray!15}AgentWorld &  0.27\\
        \bottomrule
        \end{tabular}
            
            }
        \end{minipage}
        \quad
        \hspace{8pt}
        \begin{minipage}[b]{0.30\linewidth}
        \centering
        \makeatletter\def\@captype{table}
        \setlength{\tabcolsep}{3.5mm}{
            \caption{Ablation on different model variants.}
            \label{tab:model}
            \vspace{3pt}
            \begin{tabular}{lc}
        \toprule 
        \textbf{Method} & \textbf{Avg. L2 (m)} $\downarrow$ \\
        \midrule
        Qwen2-VL-2B &  0.27\\
        \rowcolor{gray!15}Qwen2-VL-7B &  0.25\\
        Qwen2.5-VL-3B &  0.27\\
        \bottomrule
        \end{tabular}}
        \end{minipage}
        
        \quad
        \hspace{8pt}
        \begin{minipage}[b]{0.25\linewidth}
        \centering
        \makeatletter\def\@captype{table}
        \setlength{\tabcolsep}{3mm}{
            \caption{Ablation on reward of RL stage.}
            \label{tab:rl}
            \vspace{3pt}
            \begin{tabular}{lc}
        \toprule 
        \textbf{Method} & \textbf{Avg. L2 (m)} $\downarrow$ \\
        \midrule
        w/o $R_{\text{action}}$ &  0.29\\
        w/o $R_{\text{agent}}$ &  0.30\\
        \rowcolor{gray!15}w/o $R_{\text{traj}}$   &  0.32\\
        \bottomrule
        \end{tabular}}
        \end{minipage}
        
        }
\end{table}

\begin{figure}[t]
    \centering
    \includegraphics[width=0.8\linewidth]{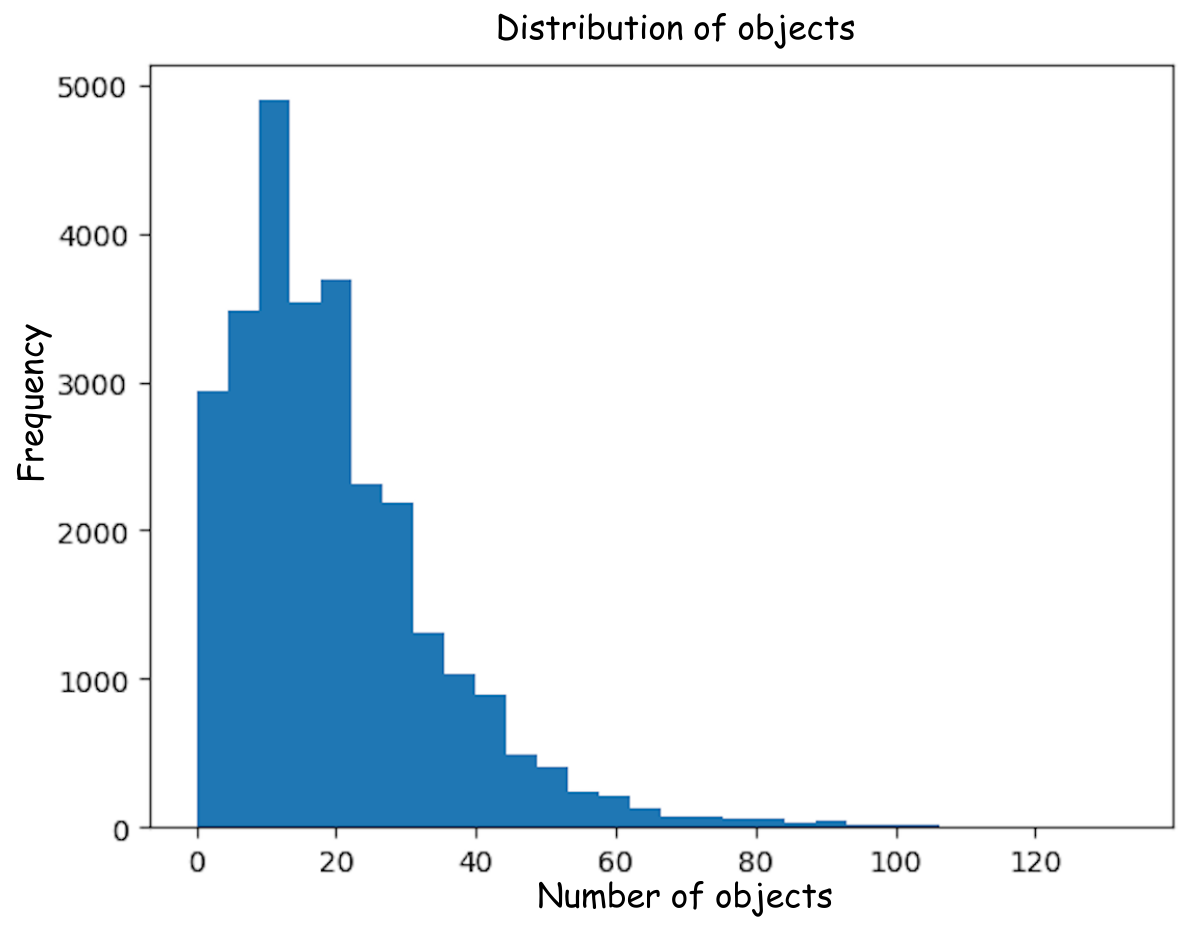}
    % \vspace{-15pt}
    \caption{Distribution of object counts.}
    \label{fig:vis_distrubution}
    % \vspace{-20pt}
\end{figure}

\begin{figure}[t]
    \centering
    \includegraphics[width=1.0\linewidth]{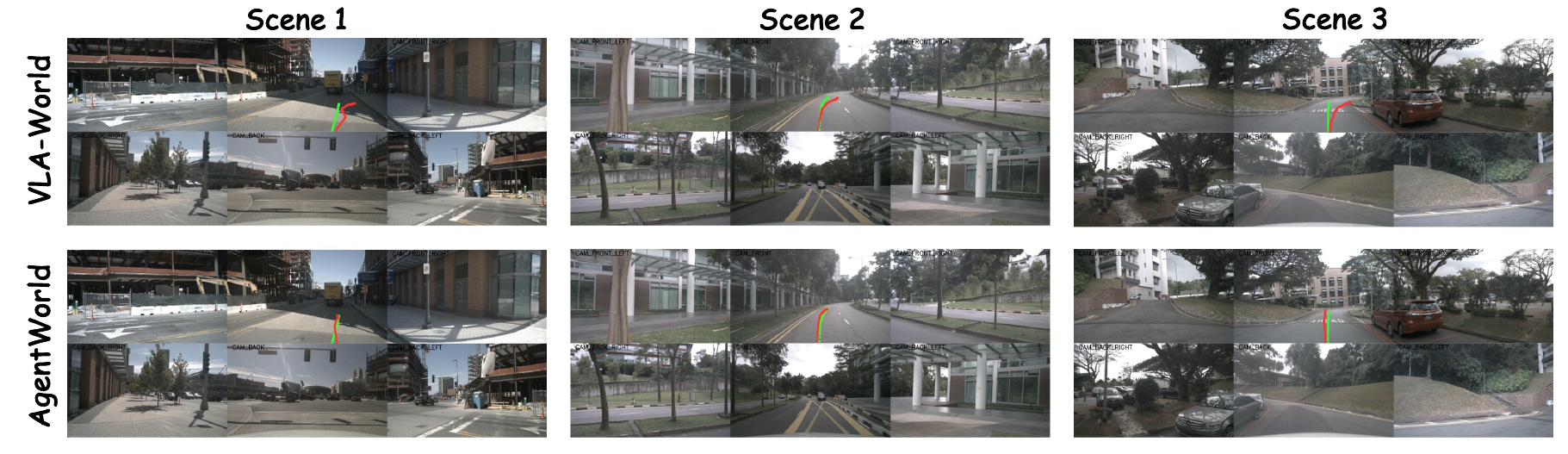}
    % \vspace{-15pt}
    \caption{Visualization of trajectories across VLA-World and our AgentWorld on the nuScenes validation dataset. The predicted and ground-truth trajectories are depicted in red and green, respectively.}
    \label{fig:vis_traj}
    % \vspace{-20pt}
\end{figure}

\subsection{More Experimental Results}
\paragraph{Ablation on agent head in SFT and RL stages.}
Tab.~\ref{tab:head} studies how the auxiliary agent head affects different training stages. Removing the agent head during SFT leads to the weakest performance, showing that explicit physical grounding is important for turning special agent tokens into usable world-state representations. Without this grounding signal, the model can still learn the response format, but the agent tokens are less reliably tied to surrounding traffic participants, which weakens downstream planning.

% Keeping the agent head during RL gives the lowest L2 error, suggesting that physical grounding can still provide useful regularization during policy optimization. However, this also keeps additional training-time scaffolding active in the RL stage. In contrast, AgentWorld removes the auxiliary head after grounding and still achieves comparable performance, indicating that the learned agent-token hidden states retain sufficient physical structure for planning. This supports the main design choice: the agent head is useful for grounding, but it does not need to remain part of the final policy interface.

The \textit{RL w head} variant achieves an Avg.~L2 error of 0.26 m, compared with 0.27 m for the default AgentWorld model. In the default model, the auxiliary agent head remains frozen during RL and is used only as a reward-side decoder to compute
agent-interaction rewards; only the MLLM policy parameters are updated through GRPO. At inference time, the auxiliary head is discarded, and the policy generates reasoning and ego waypoints using the learned agent-token representations. Thus, the auxiliary head supports physical grounding and reward computation during training without remaining part of the inference-time policy interface.

\paragraph{Ablation on different model variants.}
Tab.~\ref{tab:model} compares different MLLM backbones. The 7B variant achieves the best planning accuracy, indicating that stronger language-vision reasoning capacity improves the model’s ability to use grounded agent tokens for decision-making. This is consistent with the role of the MLLM in AgentWorld: after agent tokens provide a compact physical world-state interface, the backbone still needs sufficient reasoning capacity to identify salient agents, infer interactions, and translate them into safe trajectories. The smaller variants remain competitive, suggesting that AgentWorld is not tied to a specific backbone. The results indicate that the proposed agent-token interface is broadly compatible across MLLM scales, while larger models can further exploit this structured interface for better planning.

\paragraph{Ablation on reward in RL stage.}
Tab.~\ref{tab:rl} evaluates the contribution of different reward terms in the RL stage. Removing the trajectory reward causes the largest degradation, showing that waypoint-level feedback is the most direct signal for improving planning accuracy. This reward guides the model toward trajectories that better match expert behavior and satisfy geometric constraints.
Removing the agent-aware reward also hurts performance, confirming that interaction-level feedback is necessary for making RL consistent with the AgentWorld design. Without this term, the policy can optimize trajectory imitation while paying less attention to surrounding-agent safety, such as clearance and conflict risk. Removing the action reward produces a smaller but still clear degradation, indicating that high-level driving commands provide useful semantic guidance that stabilizes the policy before generating low-level waypoints.

Overall, the reward ablation shows that effective RL requires both trajectory-level optimization and agent-centric interaction feedback: the former improves waypoint accuracy, while the latter encourages the policy to preserve the learned agent-token reasoning interface.

\paragraph{Distribution of object counts.}
Fig.~\ref{fig:vis_distrubution} motivates the choice of the maximum agent-token number N=20. The object-count distribution is highly imbalanced: most scenes contain a moderate number of objects, while only a small fraction of samples fall into the long-tail high-density regime. This suggests that using an excessively large number of agent tokens would introduce many unused or low-relevance slots in typical scenes, increasing the reasoning cost and no-object supervision burden without proportional benefit.

Choosing N=20 provides a practical trade-off between coverage and efficiency. It covers the high-frequency region of the distribution, while the remaining crowded scenes can still be handled by the Top-K influential-agent selection strategy. Importantly, AgentWorld does not aim to exhaustively represent every annotated object; instead, it prioritizes traffic participants that are most relevant to planning, such as nearby vehicles, cut-in agents, pedestrians, and agents whose future motion may conflict with the ego trajectory. Therefore, N=20 is sufficient to preserve the key planning-relevant agents in most scenarios while keeping the agent-token world state compact.

This design also supports the motivation of AgentWorld: world modeling for driving should allocate capacity to decision-relevant agents rather than dense scene reconstruction or all visible objects. The long-tail distribution further justifies using fixed agent slots with no-object handling and influential-agent selection, rather than scaling the token count with the total number of objects.

\paragraph{Qualitative trajectory visualization.}
Fig.~\ref{fig:vis_traj} compares the predicted trajectories of VLA-World and AgentWorld against GT across representative nuScenes validation scenes. The qualitative results show that AgentWorld produces trajectories that more closely follow the ground-truth driving intent, especially in scenes requiring subtle curvature, lane following, or cautious adjustment around surrounding traffic participants.
In contrast, VLA-World shows larger deviations in several cases. This suggests that visual future imagination alone may not provide a sufficiently structured interface for planning: although future frames can be visually plausible, the policy may still fail to extract the physical and interaction cues needed to align with the expert trajectory. AgentWorld alleviates this issue by grounding the scene into agent-level tokens, allowing the policy to reason over surrounding participants directly rather than relying on dense visual features or ego-motion shortcuts.
These visualizations support the quantitative findings: AgentWorld improves planning not merely by adding more information, but by organizing scene information into a more actionable world state. The closer alignment between predicted and ground-truth trajectories indicates that agent-centric representations provide a stronger interface for translating perception and prediction into driving actions.

\subsection{Broader Impacts}
AgentWorld aims to improve autonomous driving systems by representing dynamic scenes through compact, agent-level world states. By explicitly grounding traffic participants in 3D geometry, motion, and interaction context, the proposed framework has the potential to improve the transparency and reliability of planning-oriented reasoning, especially in complex multi-agent driving scenarios. Such representations may also facilitate diagnostic analysis, since the learned agent tokens provide an interpretable interface for examining which surrounding agents influence the ego vehicle's decisions.

At the same time, autonomous driving is a safety-critical domain, and improvements in open-loop or closed-loop benchmarks should not be interpreted as sufficient evidence for real-world deployment. Failure cases may arise from imperfect perception, long-tail traffic behaviors, distribution shifts, sensor degradation, or ambiguous human intentions. In addition, agent-centric abstraction may underrepresent rare objects, vulnerable road users, or context factors that are not well captured by the predefined agent tokens. Therefore, AgentWorld should be used as a research step toward more structured and interpretable world modeling, rather than as a standalone deployable driving system. Practical deployment would require rigorous validation under diverse geographic, environmental, and social conditions, as well as careful integration with redundancy, uncertainty estimation, rule compliance, and human oversight mechanisms.

\newpage

\clearpage
\section*{NeurIPS Paper Checklist}

\begin{enumerate}

\item {\bf Claims}
    \item[] Question: Do the main claims made in the abstract and introduction accurately reflect the paper's contributions and scope?
    \item[] Answer: \answerYes{} % Replace by \answerYes{}, \answerNo{}, or \answerNA{}.
    \item[] Justification: We claim in the abstract and introduction that this paper studies advancing world vision-language-action models via agentic world states.

\item {\bf Limitations}
    \item[] Question: Does the paper discuss the limitations of the work performed by the authors?
    \item[] Answer: \answerYes{} % Replace by \answerYes{}, \answerNo{}, or \answerNA{}.
    \item[] Justification: The limitations are addressd in Section~\ref{sec:conlusion}.

\item {\bf Theory assumptions and proofs}
    \item[] Question: For each theoretical result, does the paper provide the full set of assumptions and a complete (and correct) proof?
    \item[] Answer: \answerNA{} % Replace by \answerYes{}, \answerNo{}, or \answerNA{}.
    \item[] Justification: This work does not include theoretical results or proofs, similar to previous VLA works.

    \item {\bf Experimental result reproducibility}
    \item[] Question: Does the paper fully disclose all the information needed to reproduce the main experimental results of the paper to the extent that it affects the main claims and/or conclusions of the paper (regardless of whether the code and data are provided or not)?
    \item[] Answer: \answerYes{} % Replace by \answerYes{}, \answerNo{}, or \answerNA{}.
    \item[] Justification: We provide the implementation details in the Appendix.

\item {\bf Open access to data and code}
    \item[] Question: Does the paper provide open access to the data and code, with sufficient instructions to faithfully reproduce the main experimental results, as described in supplemental material?
    \item[] Answer: \answerYes{} % Replace by \answerYes{}, \answerNo{}, or \answerNA{}.
    \item[] Justification: The data and code will be publicly available.

\item {\bf Experimental setting/details}
    \item[] Question: Does the paper specify all the training and test details (e.g., data splits, hyperparameters, how they were chosen, type of optimizer) necessary to understand the results?
    \item[] Answer: \answerYes{} % Replace by \answerYes{}, \answerNo{}, or \answerNA{}.
    \item[] Justification: We provide the training details in Section~\ref{sec:exp} and Appendix.

\item {\bf Experiment statistical significance}
    \item[] Question: Does the paper report error bars suitably and correctly defined or other appropriate information about the statistical significance of the experiments?
    \item[] Answer: \answerNA{} % Replace by \answerYes{}, \answerNo{}, or \answerNA{}.
    \item[] Justification: We set a fixed seed to ensure the reproduction of results.

\item {\bf Experiments compute resources}
    \item[] Question: For each experiment, does the paper provide sufficient information on the computer resources (type of compute workers, memory, time of execution) needed to reproduce the experiments?
    \item[] Answer: \answerYes{} % Replace by \answerYes{}, \answerNo{}, or \answerNA{}.
    \item[] Justification: We provide the compute resources in Section~\ref{sec:exp}.
    
\item {\bf Code of ethics}
    \item[] Question: Does the research conducted in the paper conform, in every respect, with the NeurIPS Code of Ethics \url{https://neurips.cc/public/EthicsGuidelines}?
    \item[] Answer: \answerYes{} % Replace by \answerYes{}, \answerNo{}, or \answerNA{}.
    \item[] Justification: We conform to the NeurIPS code of Ethics in every respect.

\item {\bf Broader impacts}
    \item[] Question: Does the paper discuss both potential positive societal impacts and negative societal impacts of the work performed?
    \item[] Answer: \answerYes{} % Replace by \answerYes{}, \answerNo{}, or \answerNA{}.
    \item[] Justification: We discuss the broader impacts of this work in the Appendix.
    
\item {\bf Safeguards}
    \item[] Question: Does the paper describe safeguards that have been put in place for responsible release of data or models that have a high risk for misuse (e.g., pre-trained language models, image generators, or scraped datasets)?
    \item[] Answer: \answerYes{} % Replace by \answerYes{}, \answerNo{}, or \answerNA{}.
    \item[] Justification: To enhance safety, we suggest incorporating fail-safe mechanisms to halt driving under uncertainty. Human oversight and deployment in controlled environments are also recommended to reduce potential misuse or unintended behaviors.

\item {\bf Licenses for existing assets}
    \item[] Question: Are the creators or original owners of assets (e.g., code, data, models), used in the paper, properly credited and are the license and terms of use explicitly mentioned and properly respected?
    \item[] Answer: \answerYes{} % Replace by \answerYes{}, \answerNo{}, or \answerNA{}.
    \item[] Justification: We properly provide this information in our paper and codes.

\item {\bf New assets}
    \item[] Question: Are new assets introduced in the paper well documented and is the documentation provided alongside the assets?
    \item[] Answer: \answerYes{} % Replace by \answerYes{}, \answerNo{}, or \answerNA{}.
    \item[] Justification: The assets introduced in the paper are well organized.

\item {\bf Crowdsourcing and research with human subjects}
    \item[] Question: For crowdsourcing experiments and research with human subjects, does the paper include the full text of instructions given to participants and screenshots, if applicable, as well as details about compensation (if any)? 
    \item[] Answer: \answerNA{} % Replace by \answerYes{}, \answerNo{}, or \answerNA{}.
    \item[] Justification: This paper does not involve crowdsourcing nor research with human subjects.

\item {\bf Institutional review board (IRB) approvals or equivalent for research with human subjects}
    \item[] Question: Does the paper describe potential risks incurred by study participants, whether such risks were disclosed to the subjects, and whether Institutional Review Board (IRB) approvals (or an equivalent approval/review based on the requirements of your country or institution) were obtained?
    \item[] Answer: \answerNA{} % Replace by \answerYes{}, \answerNo{}, or \answerNA{}.
    \item[] Justification: This paper does not involve crowdsourcing nor research with human subjects.

\item {\bf Declaration of LLM usage}
    \item[] Question: Does the paper describe the usage of LLMs if it is an important, original, or non-standard component of the core methods in this research? Note that if the LLM is used only for writing, editing, or formatting purposes and does \emph{not} impact the core methodology, scientific rigor, or originality of the research, declaration is not required.
    %this research? 
    \item[] Answer: \answerYes{} % Replace by \answerYes{}, \answerNo{}, or \answerNA{}.
    \item[] Justification: LLMs were used as the core policy for agent-grounded reasoning and trajectory planning, as described in Section 3. LLMs were also used for paper polishing. 

\end{enumerate}

\end{document}